\pdfoutput=1
\documentclass[11pt]{article}

\usepackage[preprint]{acl}

\usepackage{times}
\usepackage{latexsym}

\usepackage[T1]{fontenc}
\usepackage[utf8]{inputenc}

\usepackage{microtype}

\usepackage{inconsolata}

\usepackage{graphicx}

\usepackage[utf8]{inputenc} % allow utf-8 input
\usepackage[T1]{fontenc}    % use 8-bit T1 fonts
\usepackage{hyperref}       % hyperlinks
\usepackage{url}            % simple URL typesetting
\usepackage{booktabs}       % professional-quality tables
\usepackage{amsfonts}       % blackboard math symbols
\usepackage{nicefrac}       % compact symbols for 1/2, etc.
\usepackage{microtype}      % microtypography
\usepackage{xcolor}         % colors

\usepackage{bbding}
\definecolor{green}{HTML}{009B55}
\usepackage{microtype}
\usepackage{graphicx}
\usepackage{subfigure}
\usepackage{threeparttable}
\usepackage{booktabs} % for professional tables
\usepackage{xspace}
\usepackage[ruled,vlined]{algorithm2e}
\usepackage{macros}
\usepackage{wrapfig}
\usepackage{lipsum}
\usepackage{amsfonts}       % blackboard math symbols
\usepackage{nicefrac}       % compact symbols for 1/2, etc.
\usepackage{xcolor, colortbl}         % colors
\usepackage{multicol}
\usepackage{multirow}
\usepackage{enumitem}

\usepackage{listings}
\usepackage{pythonhighlight}

\usepackage{titletoc}
\usepackage{mathrsfs}
\usepackage{upquote}
\usepackage[linewidth=0.5pt,leftmargin=1em,rightmargin=1em,innertopmargin=5pt,
innerbottommargin=5pt,
innerrightmargin=5pt,
innerleftmargin=5pt]{mdframed}
\usepackage[toc,page,header]{appendix}

\usepackage{times}
\usepackage{latexsym}
\usepackage{tabularx}
\usepackage{multirow}
\usepackage{booktabs} % To thicken table lines
\usepackage{enumitem}
\usepackage{amsmath}

\usepackage{inconsolata}
\usepackage{graphicx}

\usepackage[utf8]{inputenc} % allow utf-8 input
\usepackage[T1]{fontenc}    % use 8-bit T1 fonts
\usepackage{hyperref}       % hyperlinks
\usepackage{url}            % simple URL typesetting
\usepackage{booktabs}       % professional-quality tables
\usepackage{amsfonts}       % blackboard math symbols
\usepackage{nicefrac}       % compact symbols for 1/2, etc.
\usepackage{microtype}      % microtypography
\usepackage{xcolor}         % colors
\usepackage{multicol}
\usepackage{transparent}
\usepackage{array}
\usepackage{bm}

\definecolor{nblue}{cmyk}{0.95,0.0,0.2,0.2}

\newcommand{\method}{\texttt{WorkForceAgent-R1}\xspace}

\usepackage{xspace}
\usepackage{cleveref}
\usepackage{colortbl}
\usepackage{listings}
\definecolor{green}{HTML}{009B55}

\usepackage{listings}
\usepackage{pythonhighlight}
\usepackage{fancyvrb}
\RecustomVerbatimCommand{\VerbatimInput}{VerbatimInput}{fontsize=\footnotesize,
  frame=single,
  framesep=0.5em, % separation between frame and text
  labelposition=topline,
}

\lstdefinestyle{prompt}{
  basicstyle=\ttfamily\small,
  breaklines=true,
  frame=none,
  keywordstyle=\bfseries,
  showstringspaces=false,
  literate={~} {$\sim$}{1},
}
\usepackage{tcolorbox}
\tcbuselibrary{listings}
\title{\method: Incentivizing Reasoning Capability in LLM-based Web Agents via Reinforcement Learning}

\author{Yuchen Zhuang$^{\spadesuit}$\thanks{~Equal contribution. Correspondence to: Hanrui Wang and Chao Zhang.}, Di Jin$^\heartsuit$\footnotemark[1], Jiaao Chen$^{\spadesuit}$, Wenqi Shi$^{\spadesuit}$, Hanrui Wang$^\heartsuit$, Chao Zhang$^{\spadesuit}$\\
$^{\spadesuit}$ Georgia Tech~~~~$^{\heartsuit}$ Independent Researcher\\
\texttt{\{yczhuang,jiaaochen,wqshi,chaozhang\}@gatech.edu}\\
\texttt{\{jindi,hanruiwang\}@alum.mit.edu}
}
\begin{document}
\maketitle
\begin{abstract}
  Large language models (LLMs)-empowered web agents enables automating complex, real-time web navigation tasks in enterprise environments. 
However, existing web agents relying on supervised fine-tuning (SFT) often struggle with generalization and robustness due to insufficient reasoning capabilities when handling the inherently dynamic nature of web interactions. 
In this study, we introduce \method{}\footnote{The code of \method{} is available at: \url{https://github.com/night-chen/WorkForceAgent-R1}.}, an LLM-based web agent trained using a rule-based R1-style reinforcement learning framework designed explicitly to enhance single-step reasoning and planning for business-oriented web navigation tasks. 
We employ a structured reward function that evaluates both adherence to output formats and correctness of actions, enabling \method{} to implicitly learn robust intermediate reasoning without explicit annotations or extensive expert demonstrations. 
Extensive experiments on the WorkArena benchmark demonstrate that \method{} substantially outperforms SFT baselines by 10.26-16.59\%, achieving competitive performance relative to proprietary LLM-based agents (\texttt{gpt-4o}) in workplace-oriented web navigation tasks.

\end{abstract}

\section{Introduction}
\label{sec:intro}
Large language models (LLMs) have emerged as powerful agents capable of executing complex tasks across diverse domains~\cite{yao2023react,liu2023agentbench}, particularly through integration with external environments or APIs~\cite{song2023llm,wang2024executable,zhuang2025hephaestus,sun2023adaplanner,liao2024reflectool,gu2024middleware,zhuang2023toolchain}. Web agents, a specialized class of LLM-powered autonomous systems, navigate and interact with websites to perform tasks ranging from partial assistance (\eg, form completion) to complete process automation (\eg, order management)~\cite{furuta2023multimodal,le2024browsergym,chezelles2024browsergym,drouin2024workarena}. Unlike structured API environments, web interfaces present unique challenges, including noisy HTML structures, dynamic content, and inconsistent element identification~\cite{he2024openwebvoyager,he2024webvoyager}, demanding robust context-aware interactions. 
Business-oriented web agents face additional complexity due to enterprise software prioritizing functionality over user experience~\cite{drouin2024workarena,xu2024theagentcompany}, resulting in complex workflows, unintuitive interfaces, and repetitive tasks. 
These challenges necessitate advanced reasoning and effective interaction strategies to reliably automate dynamic and context-rich web tasks in workplace environments.

\begin{figure}[t]
  \includegraphics[width=\linewidth]{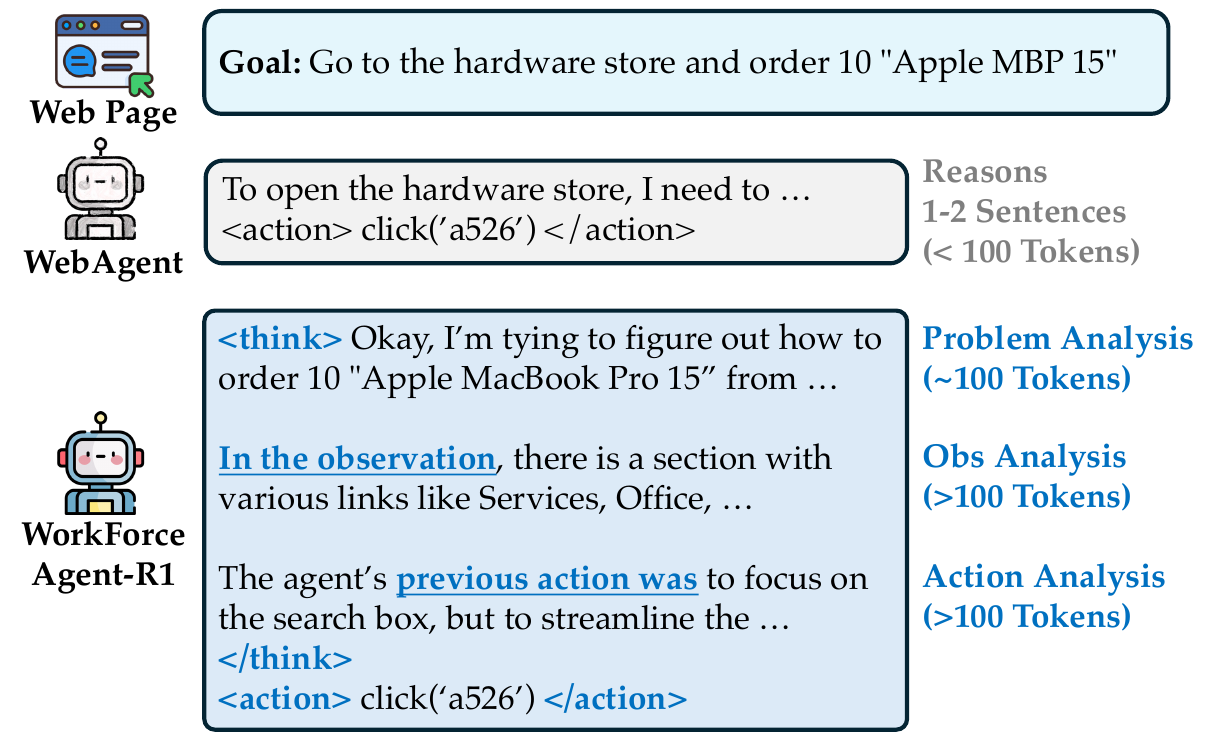}
    \caption{Example of reasoning capabilities that support service catalog in workplace with \method{}.
    }
  \label{fig:teaser}
\end{figure}

Existing web agents mainly enhance proprietary LLMs through meticulously crafted prompts or predefined workflows~\cite{he2024webvoyager,zheng2024gpt,ma2023laser}, which pose affordability and privacy concerns for workplace implementations~\cite{sun2024bbox,li2024matryoshka,Zhuang2024HYDRAMF}. 
On the other hand, significant performance gaps persist between proprietary and open-source (OSS) LLMs~\cite{drouin2024workarena,levy2024st,pan2024webcanvas,yoran2024assistantbench}, especially with inadequate post-training on complex web interactions.
Recent efforts to improve web agents via supervised fine-tuning (SFT)~\cite{lai2024autowebglm,pan2024autonomous} often narrowly optimize individual web actions without adequately addressing the inherently interactive and dynamic nature of agent-environment interactions, thus failing to guide advanced reasoning processes. 
Attempts to distill reasoning trajectories from advanced models~\cite{deng2023mind2web} frequently result in superficial ``pseudo reasoning'', encouraging models merely to intimate surface-level action patterns without genuinely internalizing underlying reasoning mechanisms~\cite{adler2024nemotron}. 

Web navigation inherently involves complex planning, challenging LLMs to execute sequential decision making, retain memory across multiple turns, and dynamically adapt to environmental feedback~\cite{he2024webvoyager,he2024openwebvoyager}. 
Observations typically consist of \emph{entire} HTML web pages, complicating multi-turn interactions between agents and their environments. 
Compared to static environments with simpler state representations, HTML data make maintaining comprehensive action-observation histories across interactions prohibitively costly and inefficient~\cite{gur2023real,ye2025realwebassist}.  
% \zc{quite redundant with the first (two) paragraphs.} 
Furthermore, the information-rich observations following each action limit the effectiveness of multi-step planning, as agents cannot reliably predict future states without direct environmental engagement. 
Simulations employing outcome-supervised reward models (ORM) to generate multi-turn trajectories~\cite{qi2024webrl} often lack sufficient precision and comprehensiveness to serve effectively as oracle guidance for complex real-world interactions. 
Consequently, developing robust web agents capable of accurate \emph{single-step planning} emerges as a computationally efficient and reliable solution for effective web automation deployments.

In this work, we introduce \method{}, a web agent trained with rule-based R1-style~\cite{guo2025deepseek} reinforcement learning (RL) that improves reasoning-driven navigation for dynamic web environments in workplaces.
We propose a progressive reward function that evaluates both format adherence and action correctness, incentivizing web agents to implicitly learn intermediate reasoning steps without relying on explicit reasoning annotations or expensive expert demonstrations. 
By decomposing multi-step web interactions into discrete planning and emphasizing explicit reasoning at each action step, \method{} effectively generalizes across diverse web interfaces while naturally balancing exploration and exploitation during training. 
Experiments on the WorkArena benchmark~\cite{drouin2024workarena} demonstrate that \method{} significantly outperforms SFT baselines by up to 16.59\%; in particular, \method{} (14B) achieves competitive performance against proprietary models (\texttt{gpt-4o} and \texttt{gpt-4.1-mini}) in workplace-oriented web navigation tasks.
We summarize our main contributions as follows:
\begin{itemize}
    \item We propose \method{}, a rule-based RL framework specifically designed to enhance \emph{single-step reasoning and planning} of LLM agents in web navigation tasks.
    \item We introduce a progressive reward function that evaluates both action correctness and adherence to structured output formats, enabling robust reasoning capabilities without explicit reasoning annotations in information-dense web environments.
    \item Extensive experiments on WorkArena demonstrate that \method{} achieves superior performance over SFT baselines competitive results relative against proprietary web agents in workplace environments.
\end{itemize}

\section{Related Works}
\label{sec:related}

\noindent \textbf{LLM-Empowered Web Agents.}
LLM-based web agents enable autonomous website navigation, dynamic content interpretation, and execution of user interactions~\cite{zhou2023webarena,deng2023mind2web,cheng2024seeclick,yao2022webshop,jang2024videowebarena}.
Early efforts primarily rely on SFT and imitation learning~\cite{lai2024autowebglm,pan2024autonomous,furuta2023multimodal,yao2022webshop,nakano2021webgpt} struggle with generalization to complex and varied web scenarios. 
More recent studies, such as WebRL~\cite{qi2024webrl} and OpenWebVoyager~\cite{he2024openwebvoyager}, have utilized RL frameworks with curriculum-driven or iterative feedback loops, achieving significant improvements in complex and dynamic web navigation tasks.
Despite these advancements, existing methods focus on optimizing \emph{individual actions} without adequately addressing the inherent complexity and interactive nature of web interactions, often relying heavily on expensive and less private-preserving proprietary models and extensive human annotations or expert demonstrations, limiting their scalability and practicality, particularly for workplace automation.

\begin{figure*}[t]
  \includegraphics[width=0.98\linewidth]{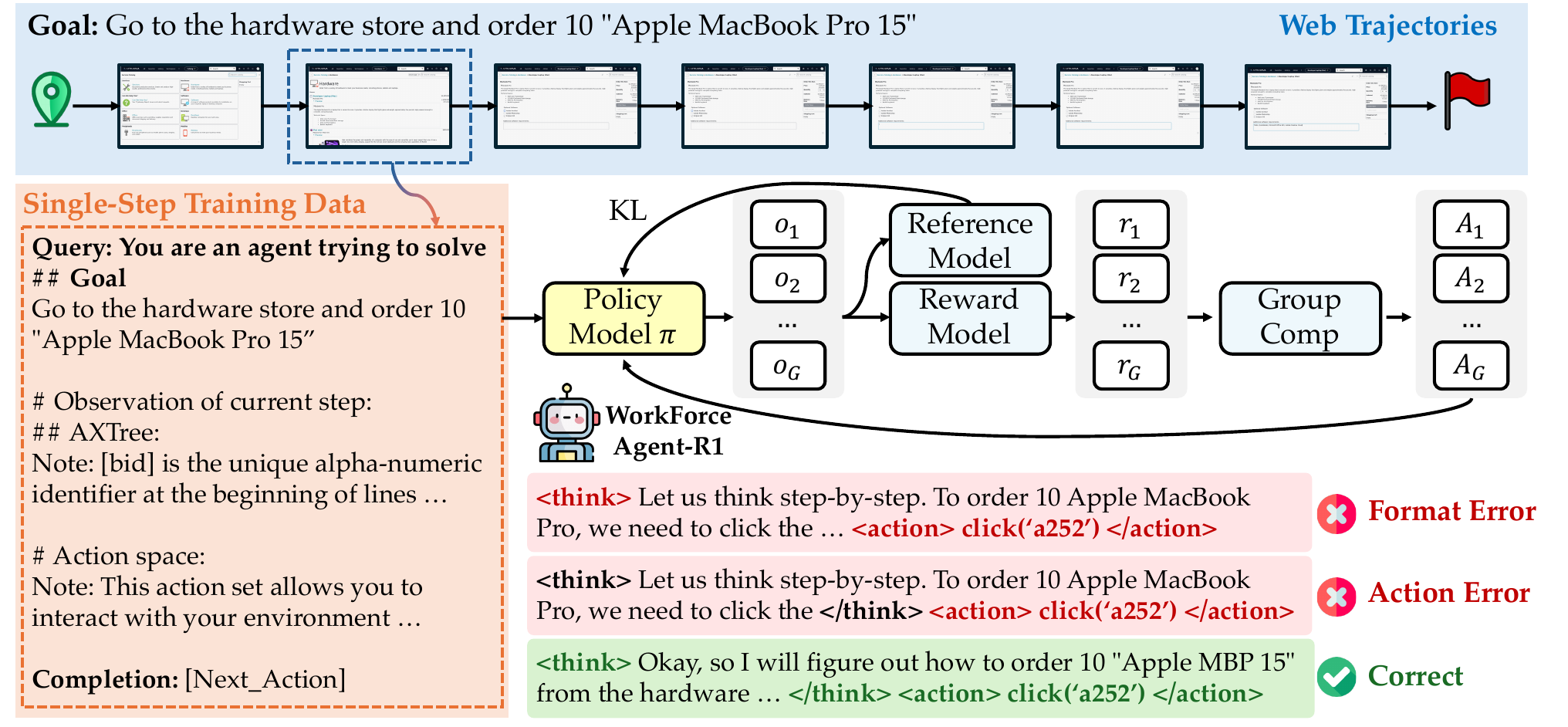}
    \caption{Overview of \method{}.
    }
  \label{fig:overview}
\end{figure*}

\noindent \textbf{RL for Improving Reasoning in LLM Agents.}  
Recent research has increasingly prioritized enhancing the reasoning capabilities of LLM agents with test-time scaling~\cite{snell2024scaling,muennighoff2025s1,adler2024nemotron}. 
DeepSeek-R1~\cite{guo2025deepseek} further demonstrates that simple rule-based RL effectively induces strong reasoning behaviors by rewarding only the correctness of the final output, thereby implicitly guiding intermediate reasoning steps.
In response, \method{} improves open-source LLM-driven web agents through R1-style RL, which facilitates robust single-step reasoning and adaptive interaction for automating web tasks in workplace environments.

\section{Preliminary}
\label{sec:prob}

Web agents performing autonomous tasks on dynamic websites face inherently complex and partially observable environments. 
Formally, we define each web navigation task as a Partially Observable Markov Decision Process (POMDP) represented by the tuple
$\mathcal{M}=(\mathcal{S},\mathcal{A},\mathcal{O},\mathcal{T},\mathcal{R})$.
Here, $\mathcal{S}$ denotes the state space comprising all possible configurations of the web environment, 
$\mathcal{A}$ is the set of permissible actions executable by the agent (\eg, clicking elements, entering text, navigating URLs), and 
$\mathcal{O}$ is the set of observations accessible to the agent. 
The transition dynamics 
$\mathcal{T:\mathcal{S}\times\mathcal{A}\to\mathcal{S}}$ specify the evolution of states based on executed actions, and $\mathcal{R}: \mathcal{S} \times \mathcal{A} \rightarrow\{0,1\}$ represents a binary reward function indicating successful task completion. 
Due to partial observability, the true state $s_t \in \mathcal{S}$ is not directly observable; instead, the agent receives observations $o_t \in \mathcal{O}$, typically represented as raw HTML web pages.

At each time step $t$, given a user instruction $I$, the LLM-based web agent must select an appropriate action $a_t \in \mathcal{A}$. Actions are defined as structured operations, each consisting of an operation name $n_i$ (\eg, ``\texttt{click}'', ``\texttt{type}'', \etc) and associated parameters $b_i$, primarily corresponding to unique element identifiers (\eg, \texttt{bid} attributes). The agent's decision-making relies upon a policy $\pi$, which maps historical context and current observations to a probability distribution over actions.
Specifically, at step $t+1$, the agent's policy $\pi$ conditions on three elements: (1) the historical context $c_t=$ $\left\{a_0, a_1, \ldots, a_t\right\}$, representing previously executed web operations, (2) the latest observation $o_t$, reflecting the most recent HTML state of the webpage, and (3) the current action set $\mathcal{A}$. The policy is formally expressed as:
\begin{equation*}
    \begin{aligned}
        \pi(a_{t+1}|c_t,o_t,\mathcal{A})=\pi(a_t|\{a_{\leq t}\},o_{t},\mathcal{A}),\ \ \text{s.t.}\ a_t\in\mathcal{A}.
    \end{aligned}
\end{equation*}
The objective is to learn a generalized policy $\pi$ that robustly addresses diverse user instructions $I$, producing sequences of action-observation pairs $\left\{\left(a_t, o_t\right)\right\}$ that consistently lead to successful task outcomes in complex web environments.

\section{\method}
\label{sec:method}

We introduce \method{}, an LLM-driven web agent specifically optimized for robust single-step reasoning in workplace-oriented web navigation tasks (Figure~\ref{fig:overview}). 
\method{} combines behavior cloning through SFT with Group Relative Policy Optimization (GRPO), utilizing structured reward signals designed to enhance both reasoning and precise action execution. 

\subsection{Data Preparation}\label{subsec:data_prep}

Previous research has extensively explored SFT for enhancing the browsing and problem-solving capabilities of LLM-based web agents~\cite{furuta2023multimodal,lai2024autowebglm,pan2024autonomous}. These approaches typically rely on datasets comprising natural language queries $Q$, paired with sequences of web operation actions and observations, denoted as $(a_0,o_0,\cdots,a_t,o_t)$.
Despite their efficacy, SFT methods frequently exhibit limited generalization due to overfitting to memorized trajectories, thereby failing to induce robust intrinsic reasoning abilities in LLMs~\cite{qi2024webrl}. 
Furthermore, collecting extensive SFT data demands significant human annotation efforts or costly API queries to advanced models (\eg, \texttt{gpt-4.1}, \texttt{o3-mini}) for expert trajectory generation.

\noindent\textbf{Data Configuration, Splitting, and Augmentation.}
We utilize the BrowserGym environment~\cite{le2024browsergym} to generate scalable web agent interactions and employ the WorkArena benchmark~\cite{drouin2024workarena} to evaluate agent performance rigorously. 
BrowserGym provides standardized interaction environments with clearly defined observation and action spaces, while WorkArena offers a remote-hosted evaluation framework comprising $33$ distinct web navigation tasks, such as form filling, knowledge base searching, and dashboard navigation. 
Each task in WorkArena is defined by a configuration file specifying target states, relevant webpage elements, and expected outcomes. To ensure data integrity and prevent information leakage, we allocate separate training and testing data configurations. 
Specifically, we reserve $10$ distinct configuration files per task, totaling $330$ training configurations. 
We further mitigate potential data leakage by leveraging advanced LLMs (\texttt{o3-mini}) to perturb the training configurations, ensuring uniqueness relative to BrowserGym and WorkArena.

\noindent\textbf{Ground-Truth Trajectories.}  
BrowserGym incorporates heuristic-based \texttt{cheat()} functions employing Playwright\footnote{\url{https://playwright.dev/}} scripts to autonomously generate oracle-like ground-truth trajectories for web interactions. 
Utilizing this functionality, we systematically sample ground-truth web navigation trajectories $(a_0,o_0,\cdots,a_T,o_T)$ for each training configuration. Due to occasional inconsistencies and redundancies in heuristic-generated actions, we enforce data standardization by removing samples with invalid actions or unsuccessful outcomes, resulting in high-quality training trajectories.

\noindent\textbf{Single-Step Reasoning.}
While recent methods~\cite{ragen,Agent-R1,qian2025toolrl} emphasize interleaving reasoning with environmental interactions, their application is constrained to simple or simulated settings. Web browsing scenarios present unique challenges: (1) observations, derived from interactions with webpages, are dynamically generated and thus complicate multi-step forward planning; (2) web observations, typically comprising extensive HTML content, impose computational and latency constraints when sequentially appended to actions. Consequently, we reformulate the original multi-step ground-truth trajectories $(a_0,o_0,\cdots,a_T,o_T)$  into single-step reasoning sequences:
\begin{equation*}
    \begin{aligned}
        \{\tau_t\}_{t=1}^T:=\{(a_0,a_1,\cdots,a_{t-1},o_{t-1})\}_{t=1}^T.
    \end{aligned}
\end{equation*}

\subsection{Thinking Template}
\begin{figure}[t]
  \includegraphics[width=0.98\linewidth]{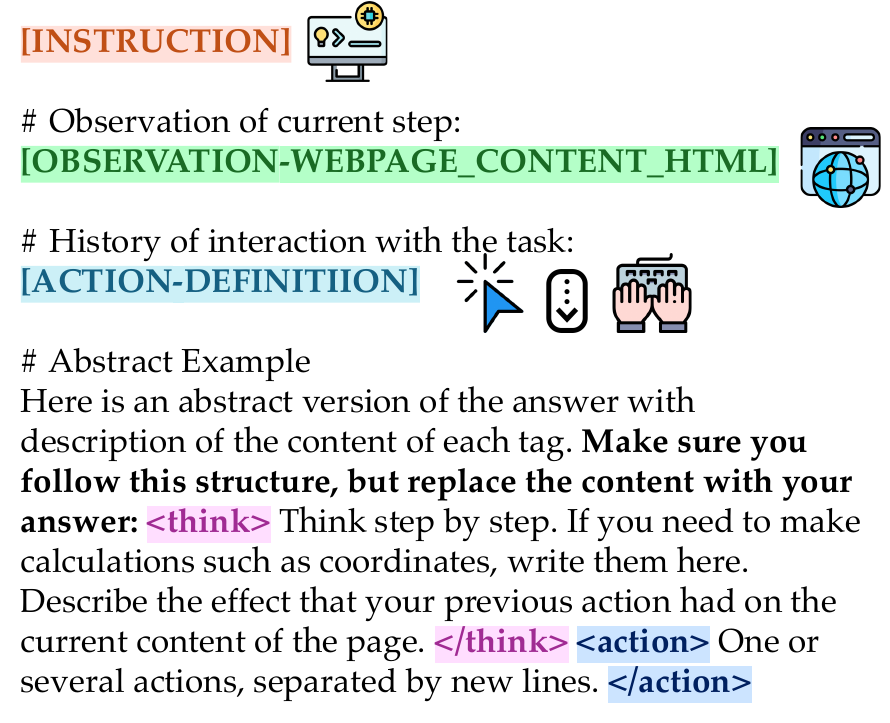}
    \caption{Example of a thinking template with an agent prompt in \method{}.
    }
  \label{fig:prompt}
\end{figure}

Following \citet{guo2025deepseek}, we introduce a structured prompting template to facilitate explicit intermediate reasoning within LLM-generated actions. This template clearly delineates intermediate reasoning with \texttt{<think>...</think>} tags and final web actions with \texttt{<action>...</action>} tags, as illustrated in Figure~\ref{fig:prompt}. Such design aims to balance structural guidance and flexibility, reducing overfitting to specific prompt patterns and promoting robust generalization across diverse web interaction scenarios~\cite{yao2023react}.

\subsection{Web Agent Training}
\subsubsection{Behavior Cloning via SFT}
We employ behavior cloning (BC) to initialize the web agent policy, leveraging SFT on action trajectories generated by heuristic methods within WorkArena. Specifically, we train the LLM to predict the correct action given the observation and context, using the standard cross-entropy loss:
\begin{equation*}
    \begin{aligned}
        \mathcal{L}_{\text{SFT}}=-\mathbb{E}_{(x,y)\sim\mathcal{D}_{\text{SFT}}}\left[\sum_{l=1}^L\log f_\theta(y_l|y_{<l},x)\right].
    \end{aligned}
\end{equation*}
where $x$ denotes the input context and observation, and $y$ represents the ground-truth action with reasoning steps. This initial SFT step enables the agent to acquire baseline capabilities for effective web interaction.

\subsubsection{Reinforcement Learning}
To endow web agents with robust reasoning capabilities, we adopt the GRPO framework~\cite{shao2024deepseekmath}. GRPO is advantageous for web agent training due to its omission of explicit value function estimation, reduced memory footprint, and precise reward-targeting mechanisms~\cite{ma2025sql}.

Formally, we decompose each web navigation task into a series of independent single-step decision problems. Specifically, at each decision step, given the current observation (webpage HTML) and historical context (previous actions), 
the agent policy at the $(m)$-th iteration generates a group of $G$ action candidates $\{c_1,c_2,\cdots,c_G\}$ from the previous policy $\pi_{\theta^{(m-1)}}$.
Each candidate action is evaluated using a carefully defined composite reward function (Section~\ref{subsubsec:reward_design}). 
GRPO calculates group-relative advantages and updates the policy parameters to preferentially select actions exhibiting higher relative performance within each group. The GRPO objective is formalized as:
\begin{equation*}
    \begin{aligned}
        &\mathcal{J}_{\text{GRPO}}(\theta)=\mathbb{E}_{x\sim \mathcal{D},\{c_i\}_{i=1}^{G}\sim \pi_{\theta^{(m-1)}}(y|x)}\\
        &\quad\Big[\frac{1}{G}\sum_{i=1}^G \Big(\min \left(r_i^{\text{ratio}},\text{clip}\left(r_i^{\text{ratio}},1-\epsilon,1+\epsilon\right)\right)A_i\\
        &\quad-\beta \mathbb{D}_{\text{KL}}\left(\pi_{\theta^{(m)}}||\pi_{\text{ref}}\right)\Big)\Big],
    \end{aligned}
\end{equation*}
where $r_i^{\text{ratio}}=\frac{\pi_{\theta^{(m)}}(c_i|x)}{\pi_{\theta^{(m-1)}}(c_i|x)}$ denotes the importance sampling ratio, quantifying the relative likelihood of candidate actions $c_i$ under updated versus previous policies $\pi_{\theta^{(m)}}$ compared to $\pi_{\theta^{(m-1)}}$;
$A_i$ represents the group-relative advantage; the clipping operator constrains policy updates; $\epsilon$ and $\beta$ are hyperparameters governing update magnitude and KL-divergence regularization, respectively; and $\pi_{\text{ref}}$ is the reference policy.

\subsubsection{Reward Function Design} \label{subsubsec:reward_design}
We design a structured and progressive reward mechanism comprising three distinct components: (1) format correctness, (2) action correctness, and (3) penalty constraints. 
This multilayered approach comprehensively guides the training of web agent toward generating accurate and contextually appropriate web interactions.

\noindent \textbf{Format Reward $R_f$.} 
The agent is incentivized to strictly adhere to a predefined structured output format, encapsulating reasoning within \texttt{<think>$\cdots$</think>} tags and actions within \texttt{<action>$\cdots$</action>} tags. 
Additionally, each generated action must be a valid operation defined by the environment. The format reward is formally defined as:
\begin{equation*}
    R_f=
    \begin{cases}
0.1, & \text{if output format is correct},\\
0, & \text{otherwise}.
\end{cases}
\end{equation*}

\noindent\textbf{Success Reward $R_s$.}
Action accuracy is crucial for successful web navigation. To evaluate correctness comprehensively, we employ exact matching against ground-truth actions, distinguishing between action types (\eg, click, fill) and associated parameters (\eg, element identifiers). Due to practical constraints in real-time environmental feedback, exact matching serves as an effective surrogate for real-world verification. The success reward is progressively structured as:
\begin{equation*}
    R_s=
    \begin{cases}
1, & \text{if action type and parameters are correct},\\
0.1, & \text{if only action type is correct},\\
0, & \text{otherwise}.
\end{cases}
\end{equation*}

\begin{table*}[t]
\centering
% \vspace{-2ex}
\fontsize{8}{10}\selectfont\setlength{\tabcolsep}{0.4em}
\caption{Main experimental results on WorkArena~\cite{drouin2024workarena}. \textbf{Bold} indicates the best performance under each task. \underline{Underline} indicates the second best. Notations are consistent across tables.}\label{tab:main}
% \resizebox{1.01\linewidth}{!}{
\begin{tabular}{@{}l|ccccccc|>{\columncolor{pink!12}}c>{\columncolor{blue!6}}c@{}}
\toprule
\textbf{Baselines ($\downarrow$) / Tasks ($\rightarrow$) }  & \textbf{Dashboard} & \textbf{Form}	& \textbf{Knowledge} & \textbf{Filter} &	\textbf{Sort}	& \textbf{Menu}	& \textbf{Service}	& \textbf{Overall} & \textbf{$\Delta$} \\ 
\midrule
\rowcolor{gray!12} \multicolumn{10}{c}{\emph{Base to Large Size Open-Source (OSS) LLMs}}  \\ \midrule
\multicolumn{9}{l}{\emph{Base Models}}  \\ \midrule
\texttt{Qwen2.5-3B-Instruct}~\cite{yang2024qwen2} &  5.00 & 1.00	& 5.00	& 0.00	& 9.17	& 1.50	& 0.00	& 2.62 & -\\
\texttt{Qwen2.5-7B-Instruct}~\cite{yang2024qwen2} & 12.50 & 10.00 & \textbf{30.00} & 0.00 & 0.00 & 20.00 & 15.60 & 9.42 & - \\
\texttt{Qwen3-8B}~\cite{qwen3} & \underline{15.00} & 10.00 & \underline{20.00} & 5.00 & 5.00 & 10.00 & 11.20 & 9.44 & - \\
\texttt{Qwen2.5-14B-Instruct}~\cite{yang2024qwen2} & 10.00 & 28.57 & \underline{20.00} & 0.00 & 7.61 & 45.56 & 48.97 & 23.79 & - \\ \midrule
\multicolumn{9}{l}{\emph{+Supervised Fine-Tuning (SFT)}}  \\ \midrule
\texttt{Qwen2.5-3B-Instruct-sft} & 10.00 & 10.00 & 12.00 & 6.00 & 13.00 & 34.00 & 67.80 & 26.59 & \textcolor{green}{(+23.97)}\\ 
\texttt{Qwen2.5-7B-Instruct-sft}	& \underline{15.00}	& 13.80	& \underline{20.00}	& 1.30	& 24.17	& 54.00	& 56.15	& 27.32 & \textcolor{green}{(+17.90)} \\ 
\texttt{Qwen2.5-14B-Instruct-sft} & \underline{15.00} & 13.20 & \underline{20.00} & 8.00 & 18.00 & 56.00 & 64.40 & 30.20 & \textcolor{green}{(+6.41)}\\ \midrule
% \texttt{Qwen2.5-7B-sft-longCoT}	& 30.00	& 20.00	& 50.00	& 20.00	& 20.00	& 40.00	& 48.80	& 30.86 & \textcolor{green}{(+21.44)} \\ \midrule
\multicolumn{9}{l}{\emph{+Reinforcement Learning (RL): Reasoning Models}}  \\ \midrule
\method \texttt{(3B)} & \textbf{20.00}	& \textbf{36.00}	& \textbf{30.00}	& 10.00	& 15.00	& 64.82	& 82.12	& 36.85 & \textcolor{green}{(+34.23)} \\
\method \texttt{(7B)}	& \textbf{20.00}	& 19.80	& \textbf{30.00}	& \underline{17.00}	& \textbf{37.00}	& \textbf{82.00}	& \underline{69.81}	& \underline{39.56} & \textcolor{green}{(+30.14)}\\
\method \texttt{(14B)} & \textbf{20.00} & \underline{30.60} & \textbf{30.00} & \textbf{25.17} & \underline{31.33} & \underline{75.00} & \textbf{89.81} & \textbf{46.79} & \textcolor{green}{(+23.00)}\\ 
\midrule 
\rowcolor{gray!12} \multicolumn{10}{c}{\emph{API-based Proprietary LLMs and XL OSS LLMs (for reference only)}}  \\\midrule 
\texttt{gpt-3.5-turbo} & 20.00 & 2.00 & 0.00 & 0.00 & 8.30 & 5.00 & 5.60 & 6.10 & - \\
\texttt{gpt-4o}~\cite{hurst2024gpt} & 62.50 & 40.00 & 80.00 & 0.00 & 10.00 & 60.00 & 77.80 & 42.65 & - \\
\texttt{gpt-4o-V}~\cite{hurst2024gpt} & 72.50 & 34.00 & 70.00 & 0.00 & 13.30 & 90.00 & 65.60 & 41.80 & - \\ 
\texttt{gpt-4.1-mini}~\cite{gpt-4-1} & 25.00 & 41.80 & 25.00 & 0.33 & 7.00 & 96.50 & 95.38 & 43.27 & -\\
\texttt{gpt-4.1}~\cite{gpt-4-1} & 50.00 & 52.40 & 60.00 & 15.17 & 9.52 & 97.99 & 80.00 & 48.19 & - \\ 
\texttt{o4-mini}~\cite{o4-mini} & 44.00 & 78.00 & 54.00 & 16.00 & 27.00 & 88.60 & 84.80 & 55.78 & - \\
\texttt{Llama3-70B}~\cite{dubey2024llama} & 37.50 & 32.00 & 30.00 & 0.00 & 1.70 & 0.00 & 26.70 & 17.90 & - \\ 
\bottomrule
\end{tabular}
% }
% \vspace{-2ex}
\end{table*}

\noindent \textbf{Penalty Reward $R_p$.}
To discourage undesirable token generation and mitigate reward hacking behaviors, particularly the generation of extraneous tokens beyond defined action boundaries, we introduce a penalty term. Specifically, tokens generated after the termination tag \texttt{</action>} invalidate the action, significantly reducing the overall reward to emphasize adherence to strict completion conditions. Formally, the penalty reward is expressed as:
\begin{equation*}
    R_p=
    \begin{cases}
-0.9, & \text{if tokens appears after \texttt{<action>}},\\
0, & \text{otherwise}.
\end{cases}
\end{equation*}
Finally, the total reward $R$ for each action candidate is the summation of the three reward components:
\begin{equation*}
    \begin{aligned}
        R=R_f+R_s+R_p.
    \end{aligned}
\end{equation*}

\section{Experiments}
\label{sec:exp}

\begin{figure*}[t]
	\centering
	\subfigure[Average reward]{
		\includegraphics[width=0.31\linewidth]{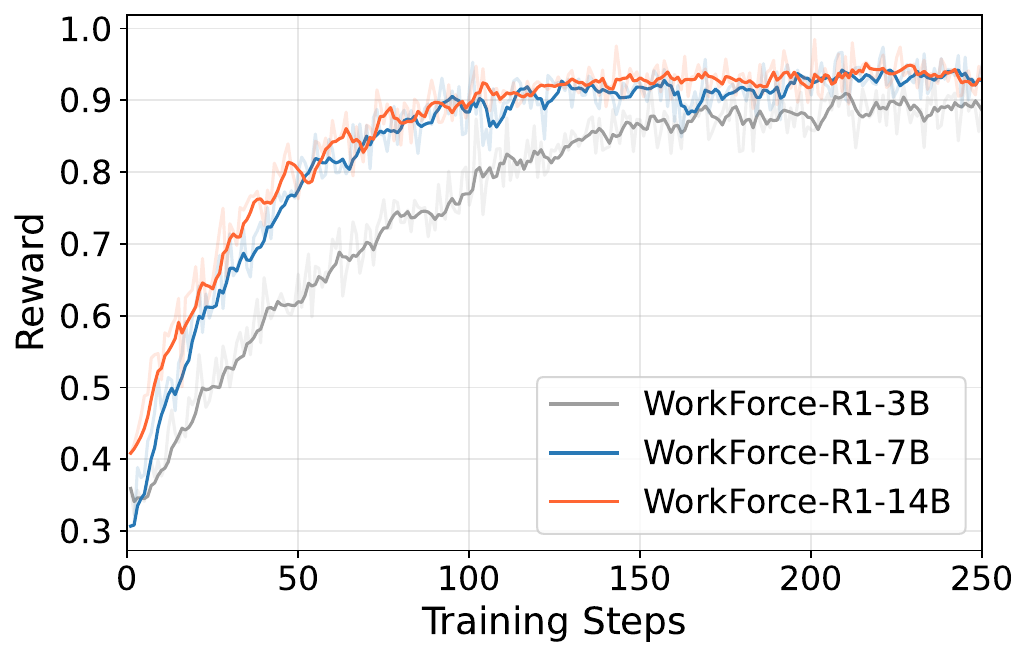}
		\label{fig:reward_curve}
	} 
     \subfigure[Average Response Length]{
		\includegraphics[width=0.31\linewidth]{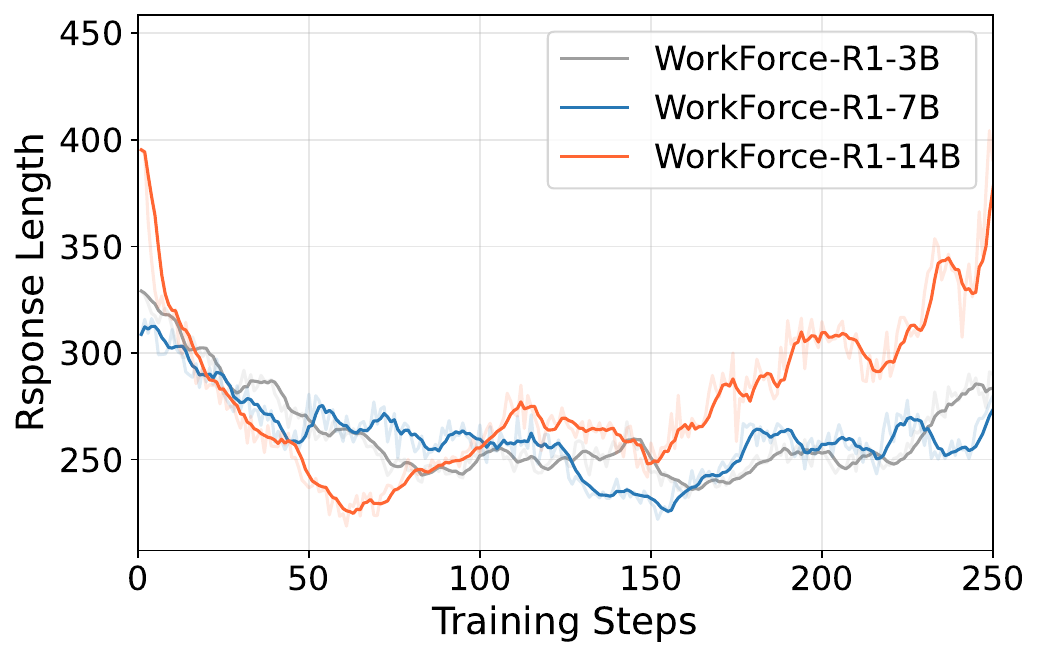}
		\label{fig:response_length_curve}
	}
     \subfigure[Validation Accuracy]{
		\includegraphics[width=0.31\linewidth]{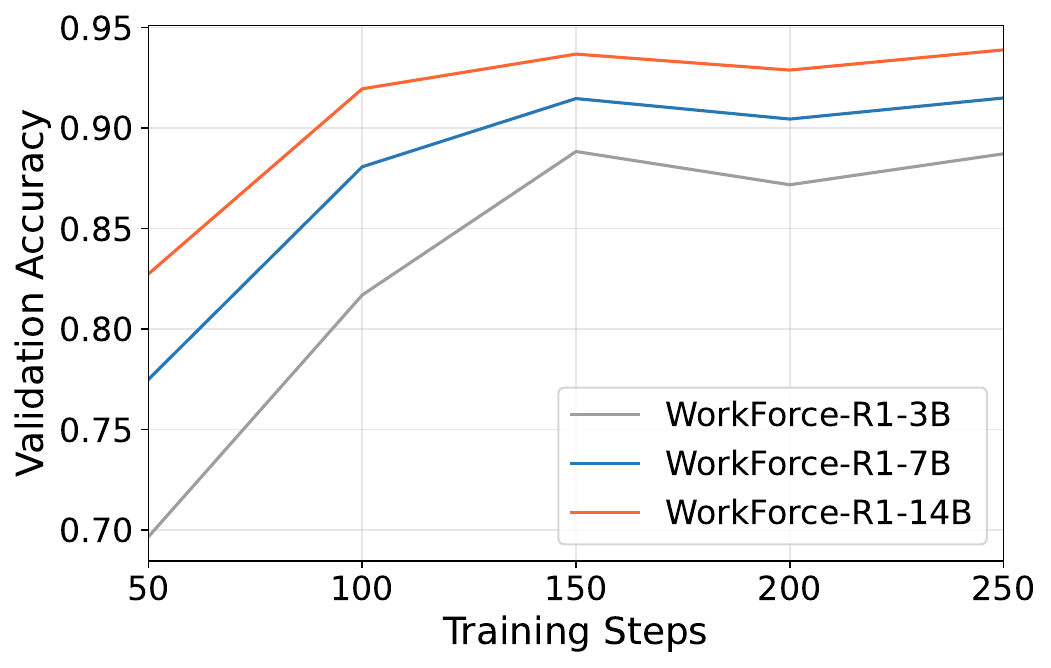}
		\label{fig:valid_acc_curve}
	}
	\caption{Learning curves across training steps.
}
\label{fig:compare}
\end{figure*}

\subsection{Experiments Setup}

\noindent \textbf{Tasks and Datasets.}
We conduct experiments on the WorkArena~\cite{drouin2024workarena} benchmark, covering 7 categories of tasks in workplaces: {dashboards}, {forms}, {knowledge} bases, list {filter}ing, list {sort}ing, {menus}, and {service}. 
Detailed descriptions are in \cref{app:data}.

\noindent \textbf{Baselines.}
We consider the following baselines for comparison: (1) API-based proprietary LLMs for reference, including \texttt{gpt-3.5-turbo}, \texttt{gpt-4o}, \texttt{gpt-4o-V}, \texttt{gpt-4.1-mini}, \texttt{gpt-4.1}, and \texttt{o4-mini};
(2) small- to XL-size OSS LLMs, including \texttt{Qwen2.5-Instruct}, \texttt{Qwen3}, and \texttt{LLaMA-3-70B}; and (3) SFT of OSS LLMs. 

\noindent \textbf{Evaluation Metrics.}
Following \citet{drouin2024workarena}, we adopt \textit{success rate (SR)} as the primary evaluation metric. 
The overall score represents the \emph{weighted average} of the success rates across various task categories, adjusted according to the number of tasks in each category.

\noindent \textbf{Backbones.} We leverage varying sizes of OSS LLMs, including \texttt{Qwen2.5-Instruct}~\cite{yang2024qwen2} (3B/7B/14B) as backbone LLMs.

\noindent \textbf{Implementation Details.} All RL experiments are conducted using the open-source framework VeRL~\cite{sheng2024hybridflow}. The SFT (warm-up) phase is executed for 1 epoch with a batch size of 32 and a learning rate of $1\times10^{-4}$ on 1,000 randomly selected samples from the training dataset. Subsequently, RL training is performed with a batch size of 128 and a learning rate of $1\times10^{-5}$. The temperature parameter during model rollout is consistently set to 0.6. Throughout training, the coefficient for the KL divergence regularization term is fixed at $\beta=1\times10^{-3}$. 
All experiments are run on 8 NVIDIA H200 GPUs, each equipped with 141GB of memory.

\subsection{Main Experiment Results}

Table~\ref{tab:main} compares \method with base models and SFT baselines. We summarize our main observations:
(1) \textbf{Superior Empirical Performance via RL:} \method{} consistently outperforms existing OSS models by an average margin of $29.13\%$ when employing similar model sizes. Remarkably, the 14B-parameter variant of \method{} surpasses the proprietary state-of-the-art model \texttt{GPT-4o} by $4.99\%$, underscoring its effectiveness in workplace-oriented web navigation tasks.
(2) \textbf{Balanced Performance Across Tasks:} Models trained via SFT exhibit significant performance imbalance, particularly evident in achieving less than $10\%$ accuracy on list-filtering tasks while nearing $60\%$ accuracy on service catalog tasks. In contrast, \method{}, benefiting from RL's enhanced generalization, achieves a more balanced performance distribution across diverse task categories.
(3) \textbf{Emergent Capabilities in Larger Models:} We observe emergent capabilities correlated with increased model sizes. Specifically, the 14B variants across all training strategies exhibit substantially improved performance over their smaller counterparts, indicating that larger parameter capacities facilitate more effective RL updates.

\subsection{Training Recipes}

\textbf{Training Dynamics.}
Figure~\ref{fig:compare} presents training dynamics of \method{}. The consistent improvements in average reward (Figure~\ref{fig:reward_curve}) and validation accuracy (Figure~\ref{fig:valid_acc_curve}) demonstrate stable learning trajectories across training steps. 
Larger models attain higher rewards at an accelerated pace compared to smaller variants. 
Notably, Figure~\ref{fig:response_length_curve} indicates that the 14B-parameter model initially reduces response length, subsequently stabilizing and then increasing to accommodate richer reasoning, suggesting that longer reasoning chains contribute positively to improved performance.

\begin{figure}[t]
  \includegraphics[width=0.98\linewidth]{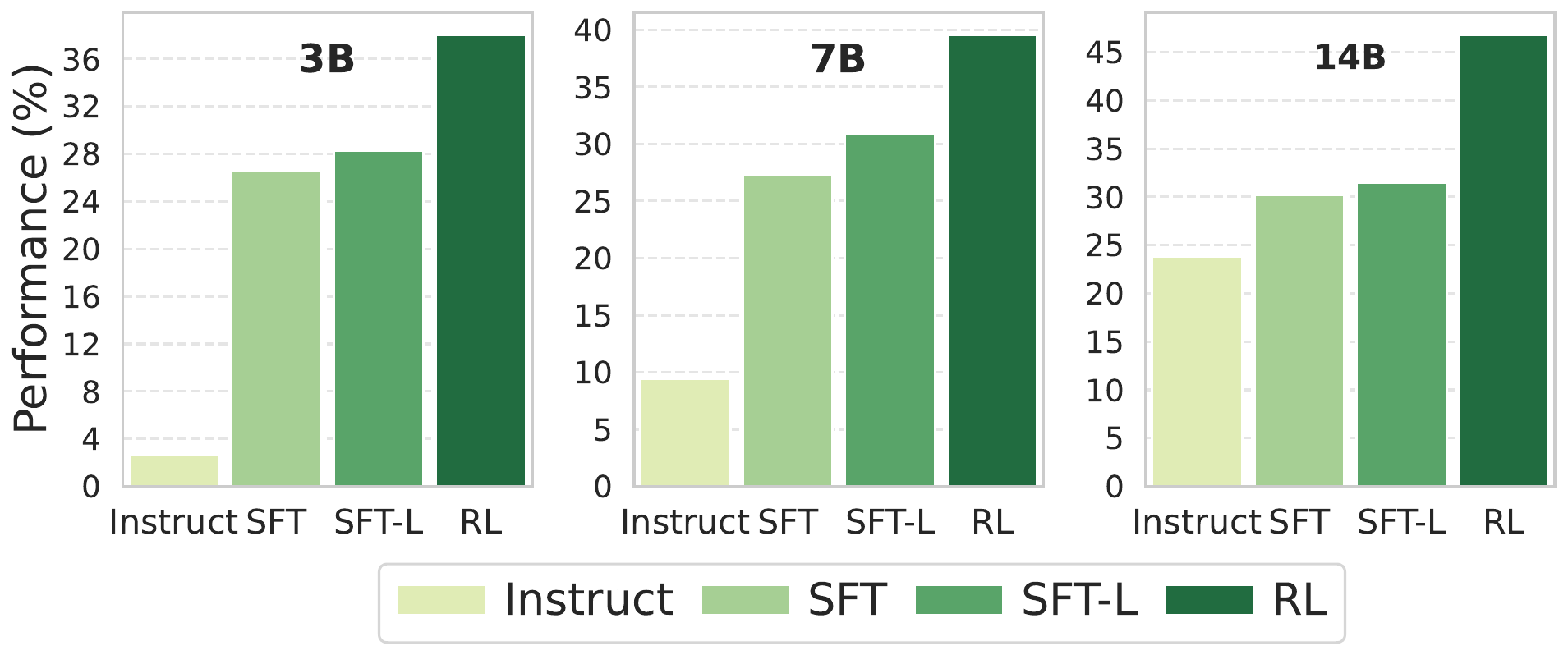}
    \caption{Performance comparison among different training strategies. ``-L'' denotes long reasoning models.
    }
  \label{fig:training}
\end{figure}

\noindent\textbf{Effect of Long Reasoning.}
Figure~\ref{fig:training} compares the effectiveness of different training methodologies. Baseline models are directly adopted from \texttt{Qwen2.5-Instruct} models, while SFT models use the \texttt{o3-mini}-annotated trajectories from BrowserGym. For SFT-L models, we employ the \texttt{deepseek-ai/DeepSeek-R1-Distill-Llama-70B} to inject long-chain reasoning explicitly. The results clearly demonstrate the advantage of incorporating explicit reasoning into training, with SFT-L substantially outperforming standard SFT.

\begin{figure}[t]
	\centering
    \subfigure[Performance]{
		\includegraphics[width=0.27\linewidth]{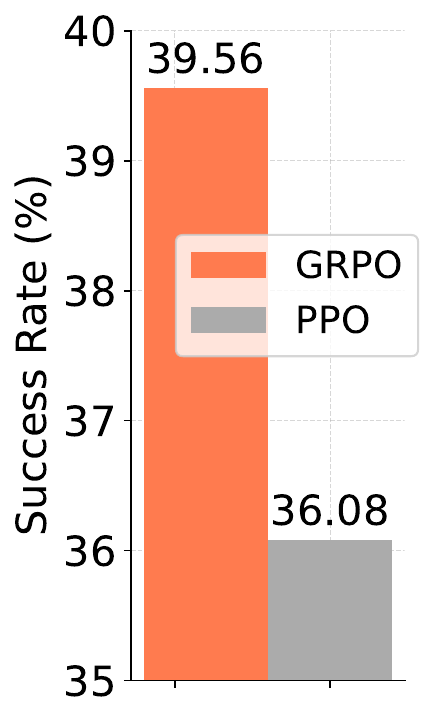}
		\label{fig:perf-ppo}
	}
    \subfigure[Average Reward]{
		\includegraphics[width=0.64\linewidth]{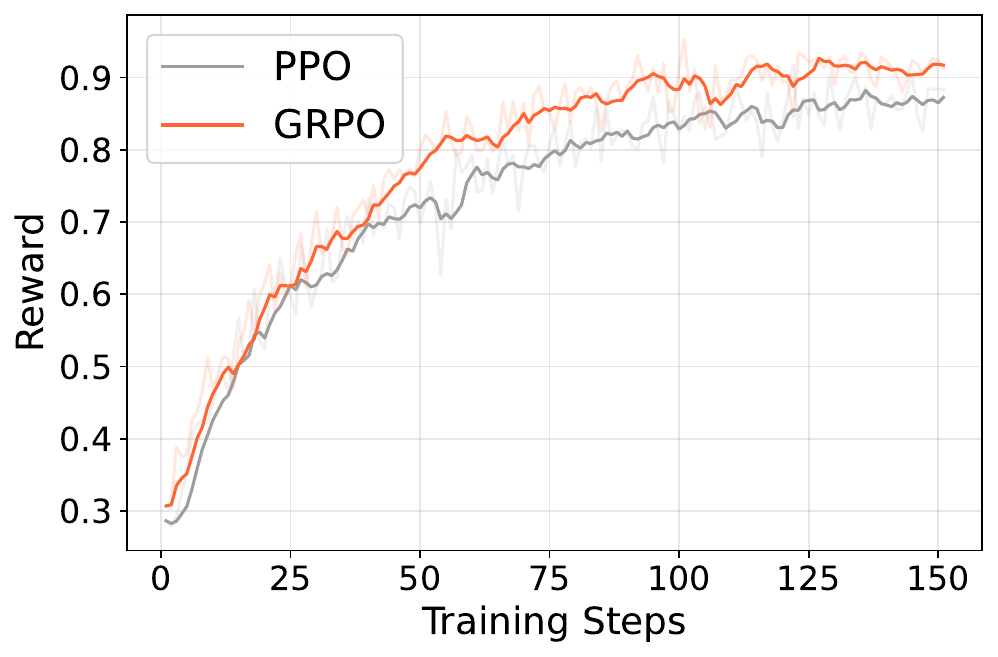}
		\label{fig:reward-ppo}
	}
	\caption{Comparison between PPO and GRPO with \texttt{Qwen2.5-7B-Instruct} as backbone LLM.
}
\label{fig:ppo}
\end{figure}

\begin{figure}[t]
	\centering
    \subfigure[Performance]{
		\includegraphics[width=0.32\linewidth]{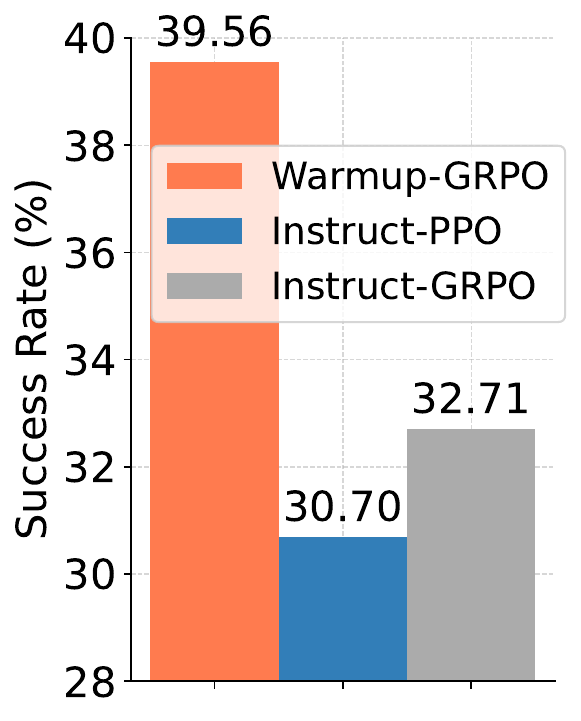}
		\label{fig:reward_ckpt-sr}
	}
    \subfigure[Average Reward]{
		\includegraphics[width=0.59\linewidth]{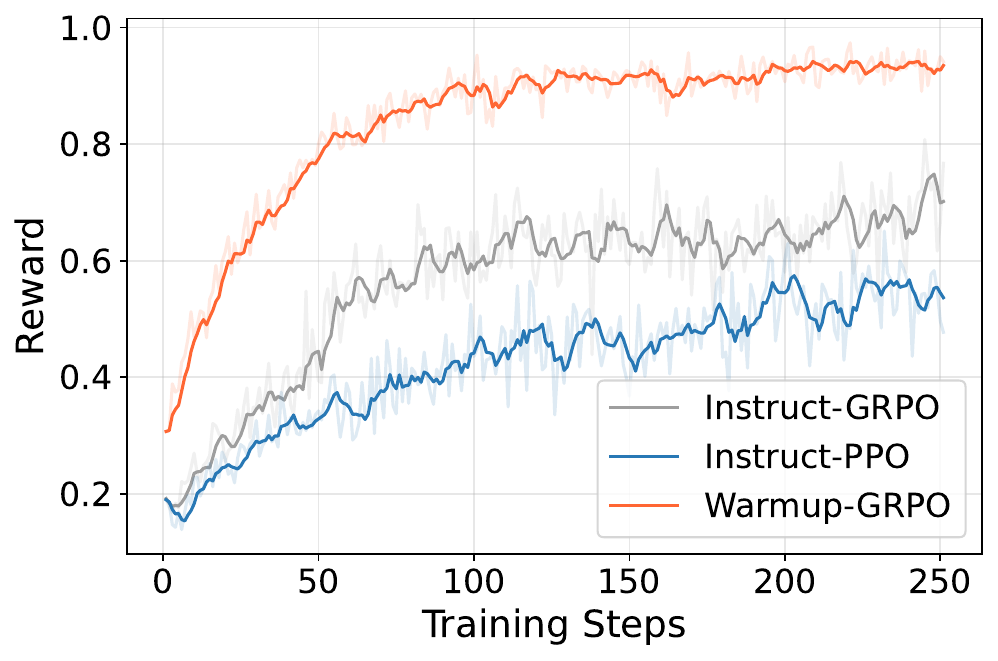}
		\label{fig:reward_ckpt-old}
	}
	\caption{Comparison between warmup model and instruct model as the initial checkpoint for RL training.
}
\label{fig:warmup}
\end{figure}

\begin{figure}[t]
	\centering
	\subfigure[Average Reward]{
		\includegraphics[width=0.45\linewidth]{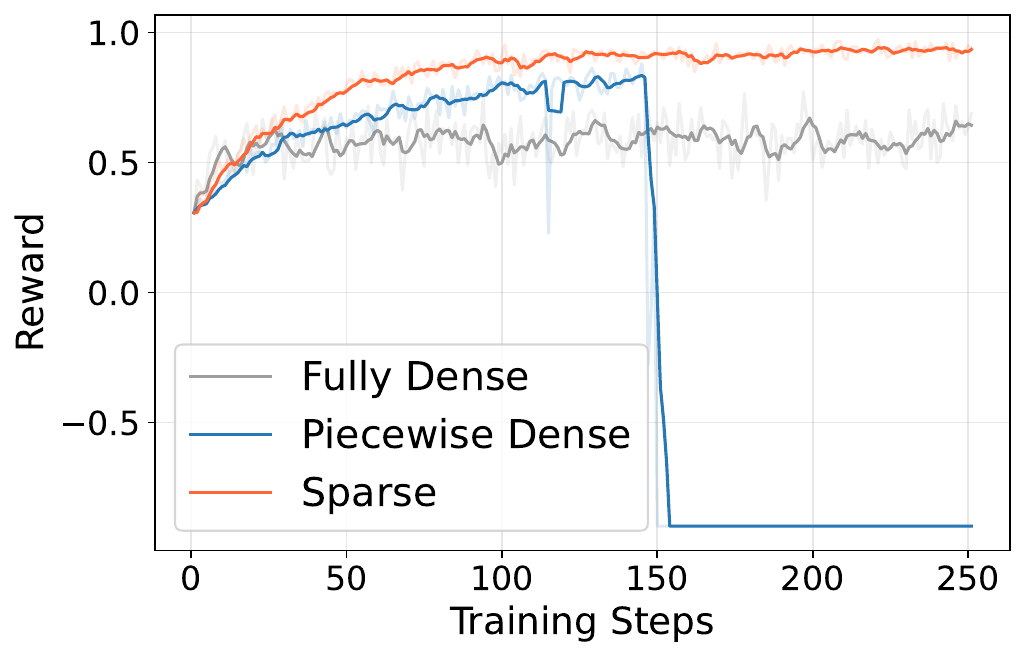}
		\label{fig:dense_reward}
	} 
     \subfigure[Average Response Length]{
		\includegraphics[width=0.45\linewidth]{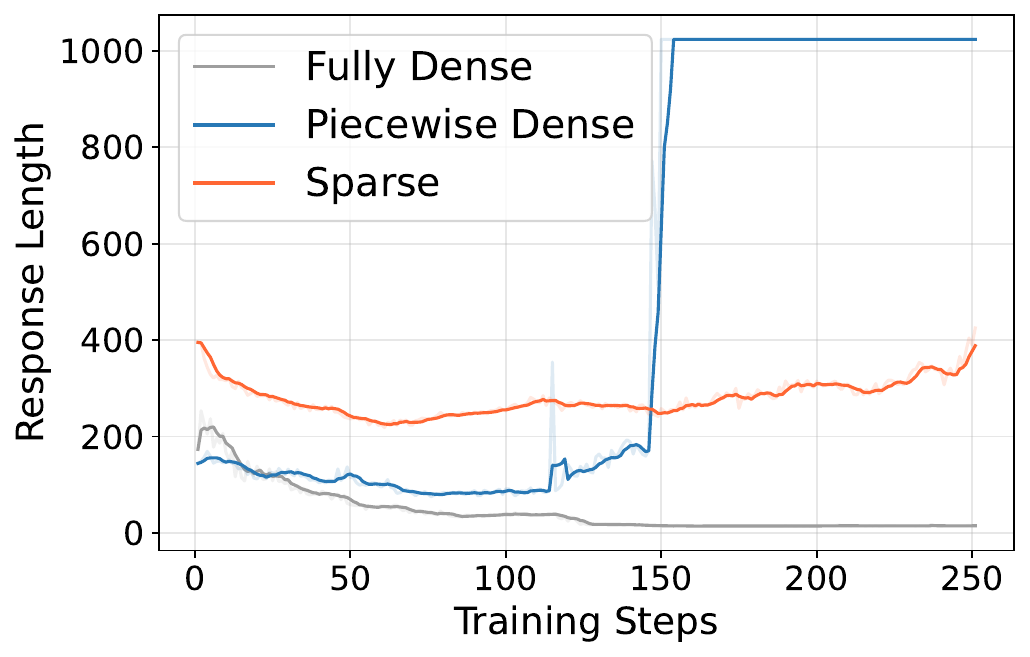}
		\label{fig:dense_reponse}
	}
	\caption{Ablation on reward granularity, comparing our ``Sparse'' loss with fine-grained reward designs, ``Fully Dense'' and ``Piecewise Dense'' rewards.
}
% \vspace{-1ex}
\label{fig:dense}
\end{figure}

\begin{table*}[t]
\centering
\fontsize{8}{10}\selectfont\setlength{\tabcolsep}{0.4em}
\caption{\texttt{LLaMA-3.1-8B} results on WorkArena~\cite{drouin2024workarena}.}\label{tab:llama}
% \resizebox{0.99\linewidth}{!}{
\begin{tabular}{@{}l|ccccccc|>{\columncolor{pink!12}}c>{\columncolor{blue!6}}c@{}}
\toprule
\textbf{Baselines ($\downarrow$) / Tasks ($\rightarrow$) }  & \textbf{Dashboard} & \textbf{Form}	& \textbf{Knowledge} & \textbf{Filter} &	\textbf{Sort}	& \textbf{Menu}	& \textbf{Service}	& \textbf{Overall} & \textbf{$\Delta$} \\ 
\midrule
\texttt{Llama-3.1-8B-Instruct}~\cite{dubey2024llama} &  5.00 & 10.00 & 20.00 & 0.00 & 0.00 & 12.00 & 18.60 & 8.57 & -\\ 
\quad + SFT & 18.00 & 12.00 & 20.00 & 4.00 & 16.00 & 36.00 & 37.80 & 20.48 & \textcolor{green}{(+11.91)} \\
\quad + SFT-L & \textbf{24.00} & \underline{16.00} & \underline{24.00} & \underline{6.00} & \underline{20.00} & \underline{41.00} & \underline{42.35} & \underline{24.56} & \textcolor{green}{(+15.99)}\\
\quad + \method & \underline{20.00} & \textbf{24.00} & \textbf{32.00} & \textbf{16.00} & \textbf{32.00} & \textbf{63.00} & \textbf{48.00} & \textbf{32.41} & \textcolor{green}{(+23.84)} \\ 
\bottomrule
\end{tabular}
% }
\end{table*}

\noindent \textbf{Effect of RL Strategies.}
Figure~\ref{fig:ppo} illustrates the comparative analysis between the GRPO and Proximal Policy Optimization (PPO) algorithms. GRPO consistently achieves higher average rewards and demonstrates more rapid convergence, confirming its effectiveness in group-based relative optimization compared to the standard PPO algorithm.

\noindent\textbf{Effect of Initial Checkpoint for RL.}
Figure~\ref{fig:warmup} evaluates the impact of initial model (\texttt{Qwen2.5-7B-Instruct}) checkpoints on RL performance. The models initialized with an additional warm-up stage of SFT on a subset of $1,000$ training samples achieve better performance and higher average rewards compared to direct instruct-model initialization. This outcome indicates the critical role of preliminary SFT in effectively conditioning models for subsequent RL.

\subsection{Ablation Study}

\noindent \textbf{Effect of Reward Design.}
We systematically investigate the influence of reward granularity on \method{} training effectiveness. We compare three distinct schemes: (1) \textbf{Sparse Reward}, used by \method{}, providing discrete positive feedback for correct formats and actions; (2) \textbf{Fully Dense Reward}, employing text similarity to assign partial credit continuously; and (3) \textbf{Piecewise Dense Reward}, applies sparse reward on function name and dense reward on parameters.
Figure~\ref{fig:dense} demonstrates that the sparse reward scheme yields superior training stability and model performance. Conversely, fully and piecewise dense rewards lead to reward-hacking, evidenced by rapid yet unstable reward increments and drastically reduced response lengths, indicating superficial pattern memorization rather than meaningful reasoning.

\noindent \textbf{Dense Rewards Cause Reward Hacking.}
Under the fully dense reward scheme, the model quickly achieves an average reward near $0.6$ but subsequently exhibits significant oscillations accompanied by a severe reduction in response length, approaching zero. Empirical analysis reveals reward hacking behavior, wherein the model repeatedly generates the most common action (\eg, \texttt{click(`a324')}) observed in prompts to consistently obtain partial rewards. Similarly, the piecewise dense reward scheme encourages reward hacking through infinite action generation following the terminating token \texttt{</action>}. Overly granular rewards incentivize superficial matching behaviors rather than genuine task-relevant reasoning.

\noindent\textbf{Effect of Backbone LLMs.}
Table~\ref{tab:llama} summarizes the outcomes, demonstrating that although \texttt{LLaMA-3.1-8B-Instruct} performs relatively lower than similarly sized Qwen-2.5 models, applying the R1-style RL consistently enhances reasoning performance across diverse backbones.

\section{Conclusion}
\label{sec:conclusion}

In this paper, we present \method{}, a rule-based RL framework designed to enhance the reasoning and planning capabilities of LLM-powered web agents, specifically tailored for workplace web navigation tasks. 
By introducing a progressive reward function that evaluates both format adherence and action accuracy, \method{} enables LLM agents to implicitly develop robust reasoning skills without requiring explicit annotations of reasoning processes. 
Extensive experiments validate that \method{} significantly surpasses SFT baselines and delivers competitive performance against state-of-the-art proprietary models. 
\method{} highlights the effectiveness of RL in addressing the inherent complexity and dynamic nature of web interactions, paving the way for more generalizable and efficient web automation solutions in enterprise environments.

\section*{Limitations}
While \method{} demonstrates substantial improvements in workplace-oriented web navigation tasks, it has several inherent limitations. Firstly, there remains a performance gap between our open-source approach and API-based commercial LLMs, primarily due to disparities in scale, data quality, and fine-tuning methodologies. Secondly, \method{} explicitly focuses on workplace web navigation, which may limit its generalizability and effectiveness in other contexts or more diverse web browsing scenarios. Lastly, the reward function design, though carefully structured, may still incentivize superficial reward hacking behaviors, underscoring the need for continued refinement of reward mechanisms to better align agent incentives with genuine task execution correctness and reasoning depth.

% \section*{Privacy and Ethical Statement}
% We strictly adhere to ethical standards and privacy considerations. Our training and evaluation datasets are constructed exclusively from synthetic, heuristic-generated trajectories within controlled benchmarking environments, avoiding the use of sensitive or private data from real users. Consequently, we do not identify that \method{} poses any potential risks associated with data privacy or user confidentiality. 

\bibliography{ref}

\begin{thebibliography}{51}
\providecommand{\natexlab}[1]{#1}

\bibitem[{Adler et~al.(2024)Adler, Agarwal, Aithal, Anh, Bhattacharya, Brundyn, Casper, Catanzaro, Clay, Cohen et~al.}]{adler2024nemotron}
Bo~Adler, Niket Agarwal, Ashwath Aithal, Dong~H Anh, Pallab Bhattacharya, Annika Brundyn, Jared Casper, Bryan Catanzaro, Sharon Clay, Jonathan Cohen, and 1 others. 2024.
\newblock Nemotron-4 340b technical report.
\newblock \emph{arXiv preprint arXiv:2406.11704}.

\bibitem[{Cheng et~al.(2024)Cheng, Sun, Chu, Xu, Li, Zhang, and Wu}]{cheng2024seeclick}
Kanzhi Cheng, Qiushi Sun, Yougang Chu, Fangzhi Xu, Yantao Li, Jianbing Zhang, and Zhiyong Wu. 2024.
\newblock Seeclick: Harnessing gui grounding for advanced visual gui agents.
\newblock \emph{arXiv preprint arXiv:2401.10935}.

\bibitem[{Chezelles et~al.(2024)Chezelles, Le~Sellier, Gasse, Lacoste, Drouin, Caccia, Boisvert, Thakkar, Marty, Assouel et~al.}]{chezelles2024browsergym}
De~Chezelles, Thibault Le~Sellier, Maxime Gasse, Alexandre Lacoste, Alexandre Drouin, Massimo Caccia, L{\'e}o Boisvert, Megh Thakkar, Tom Marty, Rim Assouel, and 1 others. 2024.
\newblock The browsergym ecosystem for web agent research.
\newblock \emph{arXiv preprint arXiv:2412.05467}.

\bibitem[{Deng et~al.(2023)Deng, Gu, Zheng, Chen, Stevens, Wang, Sun, and Su}]{deng2023mind2web}
Xiang Deng, Yu~Gu, Boyuan Zheng, Shijie Chen, Sam Stevens, Boshi Wang, Huan Sun, and Yu~Su. 2023.
\newblock Mind2web: Towards a generalist agent for the web.
\newblock \emph{Advances in Neural Information Processing Systems}, 36:28091--28114.

\bibitem[{Drouin et~al.(2024)Drouin, Gasse, Caccia, Laradji, Verme, Marty, Vazquez, Chapados, and Lacoste}]{drouin2024workarena}
Alexandre Drouin, Maxime Gasse, Massimo Caccia, Issam~H. Laradji, Manuel~Del Verme, Tom Marty, David Vazquez, Nicolas Chapados, and Alexandre Lacoste. 2024.
\newblock \href {https://openreview.net/forum?id=BRfqYrikdo} {Workarena: How capable are web agents at solving common knowledge work tasks?}
\newblock In \emph{Forty-first International Conference on Machine Learning}.

\bibitem[{Dubey et~al.(2024)Dubey, Jauhri, Pandey, Kadian, Al-Dahle, Letman, Mathur, Schelten, Yang, Fan et~al.}]{dubey2024llama}
Abhimanyu Dubey, Abhinav Jauhri, Abhinav Pandey, Abhishek Kadian, Ahmad Al-Dahle, Aiesha Letman, Akhil Mathur, Alan Schelten, Amy Yang, Angela Fan, and 1 others. 2024.
\newblock The llama 3 herd of models.
\newblock \emph{arXiv preprint arXiv:2407.21783}.

\bibitem[{Furuta et~al.(2024)Furuta, Lee, Nachum, Matsuo, Faust, Gu, and Gur}]{furuta2023multimodal}
Hiroki Furuta, Kuang-Huei Lee, Ofir Nachum, Yutaka Matsuo, Aleksandra Faust, Shixiang~Shane Gu, and Izzeddin Gur. 2024.
\newblock Multimodal web navigation with instruction-finetuned foundation models.
\newblock \emph{The Twelfth International Conference on Learning Representations}.

\bibitem[{Gu et~al.(2024)Gu, Shu, Yu, Liu, Dong, Tang, Srinivasa, Latapie, and Su}]{gu2024middleware}
Yu~Gu, Yiheng Shu, Hao Yu, Xiao Liu, Yuxiao Dong, Jie Tang, Jayanth Srinivasa, Hugo Latapie, and Yu~Su. 2024.
\newblock \href {https://doi.org/10.18653/v1/2024.emnlp-main.436} {Middleware for {LLM}s: Tools are instrumental for language agents in complex environments}.
\newblock In \emph{Proceedings of the 2024 Conference on Empirical Methods in Natural Language Processing}, pages 7646--7663, Miami, Florida, USA. Association for Computational Linguistics.

\bibitem[{Guo et~al.(2025)Guo, Yang, Zhang, Song, Zhang, Xu, Zhu, Ma, Wang, Bi et~al.}]{guo2025deepseek}
Daya Guo, Dejian Yang, Haowei Zhang, Junxiao Song, Ruoyu Zhang, Runxin Xu, Qihao Zhu, Shirong Ma, Peiyi Wang, Xiao Bi, and 1 others. 2025.
\newblock Deepseek-r1: Incentivizing reasoning capability in llms via reinforcement learning.
\newblock \emph{arXiv preprint arXiv:2501.12948}.

\bibitem[{Gur et~al.(2023)Gur, Furuta, Huang, Safdari, Matsuo, Eck, and Faust}]{gur2023real}
Izzeddin Gur, Hiroki Furuta, Austin Huang, Mustafa Safdari, Yutaka Matsuo, Douglas Eck, and Aleksandra Faust. 2023.
\newblock A real-world webagent with planning, long context understanding, and program synthesis.
\newblock \emph{arXiv preprint arXiv:2307.12856}.

\bibitem[{He et~al.(2024{\natexlab{a}})He, Yao, Ma, Yu, Dai, Zhang, Lan, and Yu}]{he2024webvoyager}
Hongliang He, Wenlin Yao, Kaixin Ma, Wenhao Yu, Yong Dai, Hongming Zhang, Zhenzhong Lan, and Dong Yu. 2024{\natexlab{a}}.
\newblock Webvoyager: Building an end-to-end web agent with large multimodal models.
\newblock \emph{arXiv preprint arXiv:2401.13919}.

\bibitem[{He et~al.(2024{\natexlab{b}})He, Yao, Ma, Yu, Zhang, Fang, Lan, and Yu}]{he2024openwebvoyager}
Hongliang He, Wenlin Yao, Kaixin Ma, Wenhao Yu, Hongming Zhang, Tianqing Fang, Zhenzhong Lan, and Dong Yu. 2024{\natexlab{b}}.
\newblock Openwebvoyager: Building multimodal web agents via iterative real-world exploration, feedback and optimization.
\newblock \emph{arXiv preprint arXiv:2410.19609}.

\bibitem[{Hurst et~al.(2024)Hurst, Lerer, Goucher, Perelman, Ramesh, Clark, Ostrow, Welihinda, Hayes, Radford et~al.}]{hurst2024gpt}
Aaron Hurst, Adam Lerer, Adam~P Goucher, Adam Perelman, Aditya Ramesh, Aidan Clark, AJ~Ostrow, Akila Welihinda, Alan Hayes, Alec Radford, and 1 others. 2024.
\newblock Gpt-4o system card.
\newblock \emph{arXiv preprint arXiv:2410.21276}.

\bibitem[{Jang et~al.(2024)Jang, Li, Zhao, Ding, Lin, Liang, Bonatti, and Koishida}]{jang2024videowebarena}
Lawrence Jang, Yinheng Li, Dan Zhao, Charles Ding, Justin Lin, Paul~Pu Liang, Rogerio Bonatti, and Kazuhito Koishida. 2024.
\newblock Videowebarena: Evaluating long context multimodal agents with video understanding web tasks.
\newblock \emph{arXiv preprint arXiv:2410.19100}.

\bibitem[{Lai et~al.(2024)Lai, Liu, Iong, Yao, Chen, Shen, Yu, Zhang, Zhang, Dong et~al.}]{lai2024autowebglm}
Hanyu Lai, Xiao Liu, Iat~Long Iong, Shuntian Yao, Yuxuan Chen, Pengbo Shen, Hao Yu, Hanchen Zhang, Xiaohan Zhang, Yuxiao Dong, and 1 others. 2024.
\newblock Autowebglm: A large language model-based web navigating agent.
\newblock In \emph{Proceedings of the 30th ACM SIGKDD Conference on Knowledge Discovery and Data Mining}, pages 5295--5306.

\bibitem[{Le~Sellier De~Chezelles et~al.(2024)Le~Sellier De~Chezelles, Gasse et~al.}]{le2024browsergym}
Thibault Le~Sellier De~Chezelles, Maxime Gasse, and 1 others. 2024.
\newblock The browsergym ecosystem for web agent research.
\newblock \emph{arXiv e-prints}, pages arXiv--2412.

\bibitem[{Levy et~al.(2024)Levy, Wiesel, Marreed, Oved, Yaeli, and Shlomov}]{levy2024st}
Ido Levy, Ben Wiesel, Sami Marreed, Alon Oved, Avi Yaeli, and Segev Shlomov. 2024.
\newblock St-webagentbench: A benchmark for evaluating safety and trustworthiness in web agents.
\newblock \emph{arXiv preprint arXiv:2410.06703}.

\bibitem[{Li et~al.(2024)Li, Zhuang, Qiang, Sun, Dai, Zhang, and Dai}]{li2024matryoshka}
Changhao Li, Yuchen Zhuang, Rushi Qiang, Haotian Sun, Hanjun Dai, Chao Zhang, and Bo~Dai. 2024.
\newblock Matryoshka: Learning to drive black-box llms with llms.
\newblock \emph{arXiv preprint arXiv:2410.20749}.

\bibitem[{Liao et~al.(2024)Liao, Jiang, Wang, and Wang}]{liao2024reflectool}
Yusheng Liao, Shuyang Jiang, Yanfeng Wang, and Yu~Wang. 2024.
\newblock Reflectool: Towards reflection-aware tool-augmented clinical agents.
\newblock \emph{arXiv preprint arXiv:2410.17657}.

\bibitem[{Liu et~al.(2023)Liu, Yu, Zhang, Xu, Lei, Lai, Gu, Ding, Men, Yang et~al.}]{liu2023agentbench}
Xiao Liu, Hao Yu, Hanchen Zhang, Yifan Xu, Xuanyu Lei, Hanyu Lai, Yu~Gu, Hangliang Ding, Kaiwen Men, Kejuan Yang, and 1 others. 2023.
\newblock Agentbench: Evaluating llms as agents.
\newblock \emph{arXiv preprint arXiv:2308.03688}.

\bibitem[{Ma et~al.(2023)Ma, Zhang, Wang, Pan, Yu, and Yu}]{ma2023laser}
Kaixin Ma, Hongming Zhang, Hongwei Wang, Xiaoman Pan, Wenhao Yu, and Dong Yu. 2023.
\newblock Laser: Llm agent with state-space exploration for web navigation.
\newblock \emph{arXiv preprint arXiv:2309.08172}.

\bibitem[{Ma et~al.(2025)Ma, Zhuang, Xu, Jiang, Chen, and Guo}]{ma2025sql}
Peixian Ma, Xialie Zhuang, Chengjin Xu, Xuhui Jiang, Ran Chen, and Jian Guo. 2025.
\newblock Sql-r1: Training natural language to sql reasoning model by reinforcement learning.
\newblock \emph{arXiv preprint arXiv:2504.08600}.

\bibitem[{Muennighoff et~al.(2025)Muennighoff, Yang, Shi, Li, Fei-Fei, Hajishirzi, Zettlemoyer, Liang, Cand{\`e}s, and Hashimoto}]{muennighoff2025s1}
Niklas Muennighoff, Zitong Yang, Weijia Shi, Xiang~Lisa Li, Li~Fei-Fei, Hannaneh Hajishirzi, Luke Zettlemoyer, Percy Liang, Emmanuel Cand{\`e}s, and Tatsunori Hashimoto. 2025.
\newblock s1: Simple test-time scaling.
\newblock \emph{arXiv preprint arXiv:2501.19393}.

\bibitem[{Nakano et~al.(2021)Nakano, Hilton, Balaji, Wu, Ouyang, Kim, Hesse, Jain, Kosaraju, Saunders et~al.}]{nakano2021webgpt}
Reiichiro Nakano, Jacob Hilton, Suchir Balaji, Jeff Wu, Long Ouyang, Christina Kim, Christopher Hesse, Shantanu Jain, Vineet Kosaraju, William Saunders, and 1 others. 2021.
\newblock Webgpt: Browser-assisted question-answering with human feedback.
\newblock \emph{arXiv preprint arXiv:2112.09332}.

\bibitem[{OpenAI(2025{\natexlab{a}})}]{gpt-4-1}
OpenAI. 2025{\natexlab{a}}.
\newblock \href {https://openai.com/index/gpt-4-1/} {Introducing gpt-4.1 in the api}.
\newblock \emph{OpenAI Blog}.

\bibitem[{OpenAI(2025{\natexlab{b}})}]{o4-mini}
OpenAI. 2025{\natexlab{b}}.
\newblock \href {https://openai.com/index/o3-o4-mini-system-card/} {Openai o3 and o4-mini system card}.
\newblock \emph{OpenAI Blog}.

\bibitem[{Ouyang et~al.(2025)Ouyang, Yan, Luo, Cheng, Liu, Liu, Yu, and Wang}]{Agent-R1}
Jie Ouyang, Ruiran Yan, Yucong Luo, Mingyue Cheng, Qi~Liu, Zirui Liu, Shuo Yu, and Daoyu Wang. 2025.
\newblock \href {https://github.com/0russwest0/Agent-R1} {Training powerful llm agents with end-to-end reinforcement learning}.

\bibitem[{Pan et~al.(2024{\natexlab{a}})Pan, Zhang, Tomlin, Zhou, Levine, and Suhr}]{pan2024autonomous}
Jiayi Pan, Yichi Zhang, Nicholas Tomlin, Yifei Zhou, Sergey Levine, and Alane Suhr. 2024{\natexlab{a}}.
\newblock Autonomous evaluation and refinement of digital agents.
\newblock \emph{arXiv preprint arXiv:2404.06474}.

\bibitem[{Pan et~al.(2024{\natexlab{b}})Pan, Kong, Zhou, Cui, Leng, Jiang, Liu, Shang, Zhou, Wu et~al.}]{pan2024webcanvas}
Yichen Pan, Dehan Kong, Sida Zhou, Cheng Cui, Yifei Leng, Bing Jiang, Hangyu Liu, Yanyi Shang, Shuyan Zhou, Tongshuang Wu, and 1 others. 2024{\natexlab{b}}.
\newblock Webcanvas: Benchmarking web agents in online environments.
\newblock \emph{arXiv preprint arXiv:2406.12373}.

\bibitem[{Qi et~al.(2025)Qi, Liu, Iong, Lai, Sun, Zhao, Yang, Yang, Sun, Yao et~al.}]{qi2024webrl}
Zehan Qi, Xiao Liu, Iat~Long Iong, Hanyu Lai, Xueqiao Sun, Wenyi Zhao, Yu~Yang, Xinyue Yang, Jiadai Sun, Shuntian Yao, and 1 others. 2025.
\newblock Webrl: Training llm web agents via self-evolving online curriculum reinforcement learning.
\newblock \emph{The Thirteenth International Conference on Learning Representations}.

\bibitem[{Qian et~al.(2025)Qian, Acikgoz, He, Wang, Chen, Hakkani-T{\"u}r, Tur, and Ji}]{qian2025toolrl}
Cheng Qian, Emre~Can Acikgoz, Qi~He, Hongru Wang, Xiusi Chen, Dilek Hakkani-T{\"u}r, Gokhan Tur, and Heng Ji. 2025.
\newblock Toolrl: Reward is all tool learning needs.
\newblock \emph{arXiv preprint arXiv:2504.13958}.

\bibitem[{Qwen(2025)}]{qwen3}
Qwen. 2025.
\newblock \href {https://qwenlm.github.io/blog/qwen3/} {Qwen3: Think deeper, act faster}.
\newblock \emph{Qwen Blog}.

\bibitem[{Shao et~al.(2024)Shao, Wang, Zhu, Xu, Song, Bi, Zhang, Zhang, Li, Wu et~al.}]{shao2024deepseekmath}
Zhihong Shao, Peiyi Wang, Qihao Zhu, Runxin Xu, Junxiao Song, Xiao Bi, Haowei Zhang, Mingchuan Zhang, YK~Li, Y~Wu, and 1 others. 2024.
\newblock Deepseekmath: Pushing the limits of mathematical reasoning in open language models.
\newblock \emph{arXiv preprint arXiv:2402.03300}.

\bibitem[{Sheng et~al.(2024)Sheng, Zhang, Ye, Wu, Zhang, Zhang, Peng, Lin, and Wu}]{sheng2024hybridflow}
Guangming Sheng, Chi Zhang, Zilingfeng Ye, Xibin Wu, Wang Zhang, Ru~Zhang, Yanghua Peng, Haibin Lin, and Chuan Wu. 2024.
\newblock Hybridflow: A flexible and efficient rlhf framework.
\newblock \emph{arXiv preprint arXiv:2409.19256}.

\bibitem[{Snell et~al.(2024)Snell, Lee, Xu, and Kumar}]{snell2024scaling}
Charlie Snell, Jaehoon Lee, Kelvin Xu, and Aviral Kumar. 2024.
\newblock Scaling llm test-time compute optimally can be more effective than scaling model parameters.
\newblock \emph{arXiv preprint arXiv:2408.03314}.

\bibitem[{Song et~al.(2023)Song, Wu, Washington, Sadler, Chao, and Su}]{song2023llm}
Chan~Hee Song, Jiaman Wu, Clayton Washington, Brian~M Sadler, Wei-Lun Chao, and Yu~Su. 2023.
\newblock Llm-planner: Few-shot grounded planning for embodied agents with large language models.
\newblock In \emph{Proceedings of the IEEE/CVF international conference on computer vision}, pages 2998--3009.

\bibitem[{Sun et~al.(2023)Sun, Zhuang, Kong, Dai, and Zhang}]{sun2023adaplanner}
Haotian Sun, Yuchen Zhuang, Lingkai Kong, Bo~Dai, and Chao Zhang. 2023.
\newblock Adaplanner: Adaptive planning from feedback with language models.
\newblock \emph{Advances in neural information processing systems}, 36:58202--58245.

\bibitem[{Sun et~al.(2024)Sun, Zhuang, Wei, Zhang, and Dai}]{sun2024bbox}
Haotian Sun, Yuchen Zhuang, Wei Wei, Chao Zhang, and Bo~Dai. 2024.
\newblock \href {https://openreview.net/forum?id=jdRIaUu3xY} {Bbox-adapter: Lightweight adapting for black-box large language models}.
\newblock In \emph{ICML}.

\bibitem[{Wang et~al.(2024)Wang, Chen, Yuan, Zhang, Li, Peng, and Ji}]{wang2024executable}
Xingyao Wang, Yangyi Chen, Lifan Yuan, Yizhe Zhang, Yunzhu Li, Hao Peng, and Heng Ji. 2024.
\newblock Executable code actions elicit better llm agents.
\newblock In \emph{Forty-first International Conference on Machine Learning}.

\bibitem[{Wang et~al.(2025)Wang, Wang, Wang et~al.}]{ragen}
Zihan Wang, Kangrui Wang, Qineng Wang, and 1 others. 2025.
\newblock \href {https://arxiv.org/abs/2504.20073} {Ragen: Understanding self-evolution in llm agents via multi-turn reinforcement learning}.
\newblock \emph{Preprint}, arXiv:2504.20073.

\bibitem[{Xu et~al.(2024)Xu, Song, Li, Tang, Jain, Bao, Wang, Zhou, Guo, Cao et~al.}]{xu2024theagentcompany}
Frank~F Xu, Yufan Song, Boxuan Li, Yuxuan Tang, Kritanjali Jain, Mengxue Bao, Zora~Z Wang, Xuhui Zhou, Zhitong Guo, Murong Cao, and 1 others. 2024.
\newblock Theagentcompany: benchmarking llm agents on consequential real world tasks.
\newblock \emph{arXiv preprint arXiv:2412.14161}.

\bibitem[{Yang et~al.(2024)Yang, Yang, Zhang, Hui, Zheng, Yu, Li, Liu, Huang, Wei et~al.}]{yang2024qwen2}
An~Yang, Baosong Yang, Beichen Zhang, Binyuan Hui, Bo~Zheng, Bowen Yu, Chengyuan Li, Dayiheng Liu, Fei Huang, Haoran Wei, and 1 others. 2024.
\newblock Qwen2. 5 technical report.
\newblock \emph{arXiv preprint arXiv:2412.15115}.

\bibitem[{Yao et~al.(2022)Yao, Chen, Yang, and Narasimhan}]{yao2022webshop}
Shunyu Yao, Howard Chen, John Yang, and Karthik Narasimhan. 2022.
\newblock Webshop: Towards scalable real-world web interaction with grounded language agents.
\newblock \emph{Advances in Neural Information Processing Systems}, 35:20744--20757.

\bibitem[{Yao et~al.(2023)Yao, Zhao, Yu, Du, Shafran, Narasimhan, and Cao}]{yao2023react}
Shunyu Yao, Jeffrey Zhao, Dian Yu, Nan Du, Izhak Shafran, Karthik~R Narasimhan, and Yuan Cao. 2023.
\newblock \href {https://openreview.net/forum?id=WE_vluYUL-X} {React: Synergizing reasoning and acting in language models}.
\newblock In \emph{The Eleventh International Conference on Learning Representations}.

\bibitem[{Ye et~al.(2025)Ye, Shi, Shih, Yun, Roosta, and Shu}]{ye2025realwebassist}
Suyu Ye, Haojun Shi, Darren Shih, Hyokun Yun, Tanya Roosta, and Tianmin Shu. 2025.
\newblock Realwebassist: A benchmark for long-horizon web assistance with real-world users.
\newblock \emph{arXiv preprint arXiv:2504.10445}.

\bibitem[{Yoran et~al.(2024)Yoran, Amouyal, Malaviya, Bogin, Press, and Berant}]{yoran2024assistantbench}
Ori Yoran, Samuel~Joseph Amouyal, Chaitanya Malaviya, Ben Bogin, Ofir Press, and Jonathan Berant. 2024.
\newblock Assistantbench: Can web agents solve realistic and time-consuming tasks?
\newblock \emph{arXiv preprint arXiv:2407.15711}.

\bibitem[{Zheng et~al.(2024)Zheng, Gou, Kil, Sun, and Su}]{zheng2024gpt}
Boyuan Zheng, Boyu Gou, Jihyung Kil, Huan Sun, and Yu~Su. 2024.
\newblock Gpt-4v (ision) is a generalist web agent, if grounded.
\newblock \emph{arXiv preprint arXiv:2401.01614}.

\bibitem[{Zhou et~al.(2023)Zhou, Xu, Zhu, Zhou, Lo, Sridhar, Cheng, Ou, Bisk, Fried et~al.}]{zhou2023webarena}
Shuyan Zhou, Frank~F Xu, Hao Zhu, Xuhui Zhou, Robert Lo, Abishek Sridhar, Xianyi Cheng, Tianyue Ou, Yonatan Bisk, Daniel Fried, and 1 others. 2023.
\newblock Webarena: A realistic web environment for building autonomous agents.
\newblock \emph{arXiv preprint arXiv:2307.13854}.

\bibitem[{Zhuang et~al.(2024{\natexlab{a}})Zhuang, Chen, Yu, Mitra, Bursztyn, Rossi, Sarkhel, and Zhang}]{zhuang2023toolchain}
Yuchen Zhuang, Xiang Chen, Tong Yu, Saayan Mitra, Victor Bursztyn, Ryan~A. Rossi, Somdeb Sarkhel, and Chao Zhang. 2024{\natexlab{a}}.
\newblock \href {https://openreview.net/forum?id=B6pQxqUcT8} {Toolchain*: Efficient action space navigation in large language models with a* search}.
\newblock In \emph{The Twelfth International Conference on Learning Representations}.

\bibitem[{Zhuang et~al.(2024{\natexlab{b}})Zhuang, Sun, Yu, Qiang, Wang, Zhang, and Dai}]{Zhuang2024HYDRAMF}
Yuchen Zhuang, Haotian Sun, Yue Yu, Rushi Qiang, Qifan Wang, Chao Zhang, and Bo~Dai. 2024{\natexlab{b}}.
\newblock \href {https://openreview.net/forum?id=CKgNgKmHYp} {{HYDRA}: Model factorization framework for black-box {LLM} personalization}.
\newblock In \emph{The Thirty-eighth Annual Conference on Neural Information Processing Systems}.

\bibitem[{Zhuang et~al.(2025)Zhuang, Yang, Jiang et~al.}]{zhuang2025hephaestus}
Yuchen Zhuang, Jingfeng Yang, Haoming Jiang, and 1 others. 2025.
\newblock Hephaestus: Improving fundamental agent capabilities of large language models through continual pre-training.
\newblock \emph{arXiv preprint arXiv:2502.06589}.

\end{thebibliography}

\appendix
\section{Task and Data Details}
\label{app:data}

WorkArena~\cite{drouin2024workarena} is a benchmark designed to evaluate the capabilities of web agents performing common knowledge-worker tasks within enterprise software environments. Specifically, it comprises 33 diverse tasks derived from interactions on the widely-used ServiceNow platform, generating a total of $19,912$ unique task instances. Each task explicitly states a natural-language goal, ensuring clarity and consistency. The tasks span seven main categories, capturing representative workplace activities and workflows typically encountered by knowledge workers:
\begin{itemize}
    \item \textbf{Lists (12 tasks, 6,900 instances)}: Tasks involve \textbf{filtering} and \textbf{sorting} data presented in tabular forms. Agents must construct complex filters with multiple conditional clauses or sort lists according to specified criteria. These operations require precise interaction with UI elements such as hidden menus and filter forms.
	\item \textbf{Forms (5 tasks, 5,000 instances)}: These tasks involve creating new entries through forms that vary significantly in complexity. Agents must accurately complete fields, manage dynamic auto-completion fields, navigate through hidden tabs, and handle date pickers. Task completion is validated via querying backend databases.
	\item \textbf{Knowledge Bases (1 task, 1,000 instances)}: Agents search through an enterprise knowledge base to retrieve specific answers to clearly stated questions. This requires keyword-based searches followed by careful navigation and textual extraction from multiple result pages. Validation checks whether the returned information matches the accepted formats.
    \item \textbf{Service Catalogs (9 tasks, 3,550 instances)}:
Tasks require agents to navigate enterprise product catalogs, select items with specific configurations, and complete order forms. Validation involves checking that orders placed by agents include the correct items with precise specifications.
	\item \textbf{Dashboards (4 tasks, 1,862 instances)}:
Tasks focus on data retrieval from visual dashboards, requiring the agent to interpret numerical data from graphical charts and possibly perform simple reasoning tasks (e.g., identifying minimum or maximum values). Validation ensures the correctness of the extracted numeric values and labels.
	\item \textbf{Menus (2 tasks, 1,600 instances)}:
Tasks require agents to navigate complex hierarchical menus or impersonate different user profiles to achieve specific goals. Validation includes confirming successful navigation to target locations or verifying user impersonation.

\end{itemize}

Each task includes carefully constructed validation functions, which provide immediate feedback by identifying errors, such as incorrectly filled fields or invalid selections. Additionally, tasks contain oracle functions implemented using Playwright browser automation scripts. These scripts not only demonstrate task feasibility but also supply ground-truth solutions for training and benchmarking purposes.
The ServiceNow platform introduces unique challenges through its dynamic UI elements, non-standard and proprietary HTML implementations, and large-scale DOM structures, often ranging from $40$k to $500$k tokens after cleaning. Consequently, WorkArena tasks demand sophisticated context-awareness and robust generalization capabilities from the web agents.
\section{Prompt Details}
\label{app:prompt}

\subsection{Qwen-Style Prompt Template}
\begin{tcolorbox}[
    colback=gray!10,          % background color
    colframe=black!50,        % border color
    title=\bfseries Qwen-Style Prompt Template,
    fonttitle=\small\sffamily,
    sharp corners,            % remove rounding if you like
    boxrule=0.8pt,
    top=1mm, bottom=1mm, left=1mm, right=1mm
]
\begin{lstlisting}[style=prompt]
<|im_start|>system
You are a helpful assistant. You first thinks about the reasoning process in the mind and then provides the user with the answer.<|im_end|>
<|im_start|>user
(prompt)<|im_end|>
<|im_start|>assistant
Let me solve this step by step.
<think>
\end{lstlisting}
\end{tcolorbox}

\subsection{WorkArena Prompt Template -- Instruction}
\begin{tcolorbox}[
    colback=gray!10,          % background color
    colframe=black!50,        % border color
    title=\bfseries Instruction Templates,
    fonttitle=\small\sffamily,
    sharp corners,            % remove rounding if you like
    boxrule=0.8pt,
    top=1mm, bottom=1mm, left=1mm, right=1mm
]
\begin{lstlisting}[style=prompt]
You are an agent trying to solve a web task based on the content of the page and
user instructions. You can interact with the page and explore, and send messages to the user. Each time you
submit an action it will be sent to the browser and you will receive a new page.
# Instructions
Review the current state of the page and all other information to find the best
possible next action to accomplish your goal. Your answer will be interpreted
and executed by a program, make sure to follow the formatting instructions.

## Goal:
(User Query)
\end{lstlisting}
\end{tcolorbox}

\subsection{WorkArena Prompt Template -- Observation}
\begin{tcolorbox}[
    width=\linewidth,
    colback=gray!10,          % background color
    colframe=black!50,        % border color
    title=\bfseries Observation Templates,
    fonttitle=\small\sffamily,
    sharp corners,            % remove rounding if you like
    boxrule=0.8pt,
    top=1mm, bottom=1mm, left=1mm, right=1mm
]
\begin{lstlisting}[style=prompt]
# Observation of current step:

## AXTree:
Note: [bid] is the unique alpha-numeric identifier at the beginning of lines for each element in the AXTree. Always use bid to refer to elements in your actions.

Note: You can only interact with visible elements. If the "visible" tag is not
present, the element is not visible on the page.

RootWebArea 'Classic | Unified Navigation App | ServiceNow'
[47] generic, live='assertive', atomic, relevant='additions text'
[48] generic, live='polite', atomic, relevant='additions text'
[53] generic, live='polite', atomic, relevant='all'
[56] navigation 'Global skip links'
(Web Page AXTree)
\end{lstlisting}
\end{tcolorbox}

\subsection{WorkArena Prompt Template -- Action }
\begin{tcolorbox}[
    colback=gray!10,          % background color
    colframe=black!50,        % border color
    title=\bfseries Observation Templates,
    fonttitle=\small\sffamily,
    sharp corners,            % remove rounding if you like
    boxrule=0.8pt,
    top=1mm, bottom=1mm, left=1mm, right=1mm
]
\begin{lstlisting}[style=prompt]
# History of interaction with the task:
(Action History)

# Action space:
(Action Definition)

(Notes)

(Example)
\end{lstlisting}
\end{tcolorbox}
\section{Case Study}

\begin{figure*}[t]
  \includegraphics[width=\linewidth]{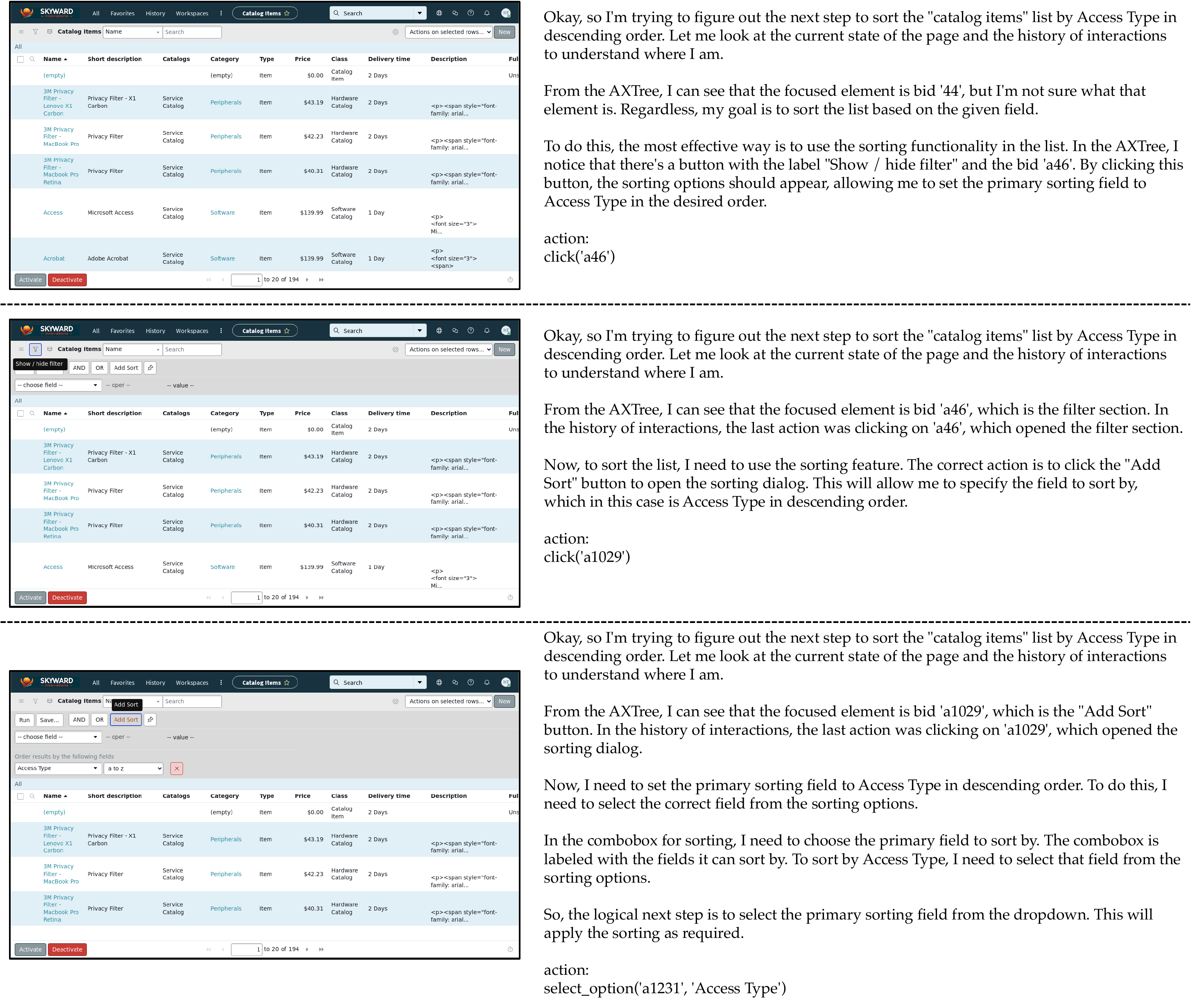}
    \caption{Case study of successful steps 1-3 of \method solving a task of \texttt{List-Sorting}.
    }
  \label{fig:case-pt1}
\end{figure*}

\begin{figure*}[t]
  \includegraphics[width=\linewidth]{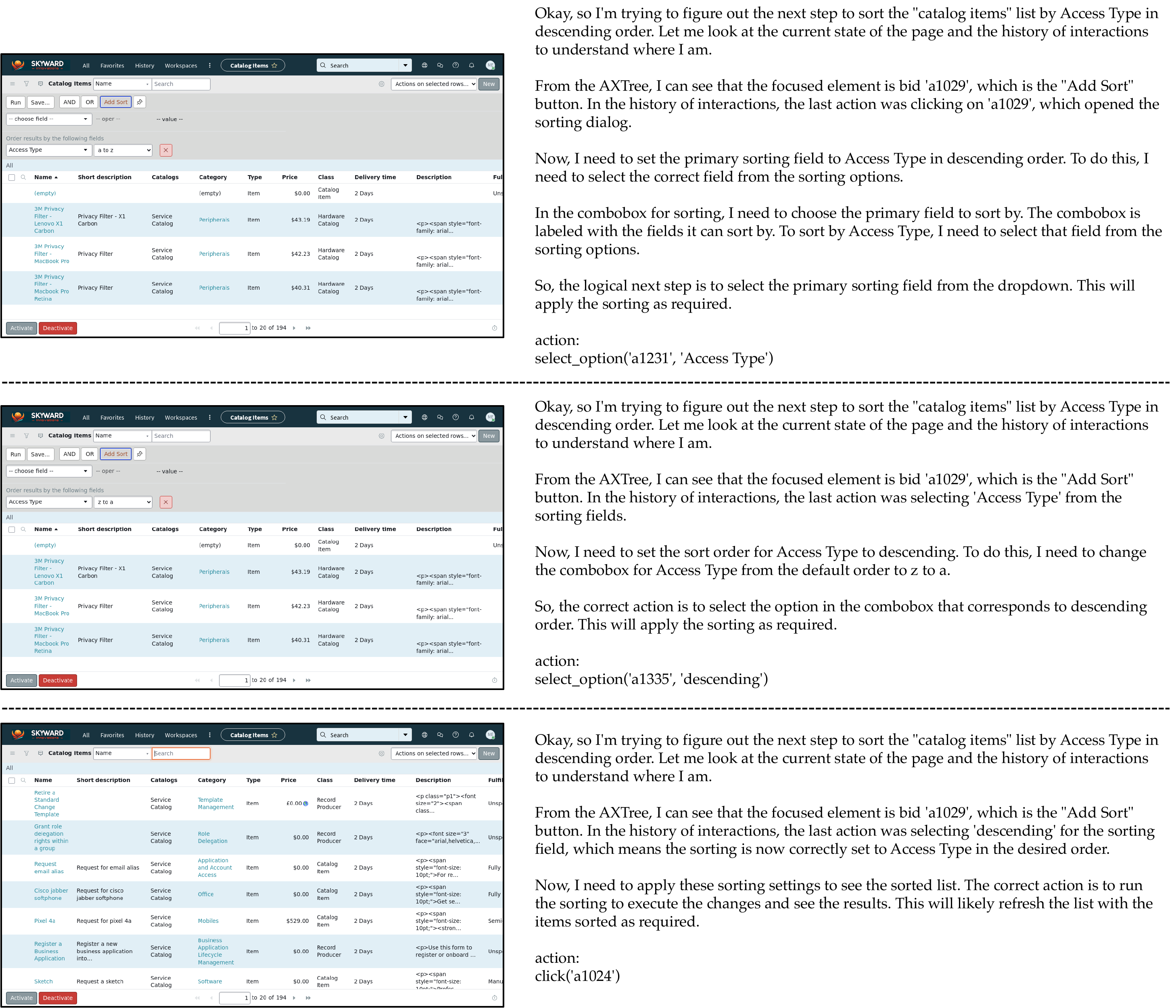}
    \caption{Case study of successful steps 4-6 of \method solving a task of \texttt{List-Sorting}.
    }
  \label{fig:case-pt2}
\end{figure*}

\begin{figure*}[t]
\centering
  \includegraphics[width=\linewidth]{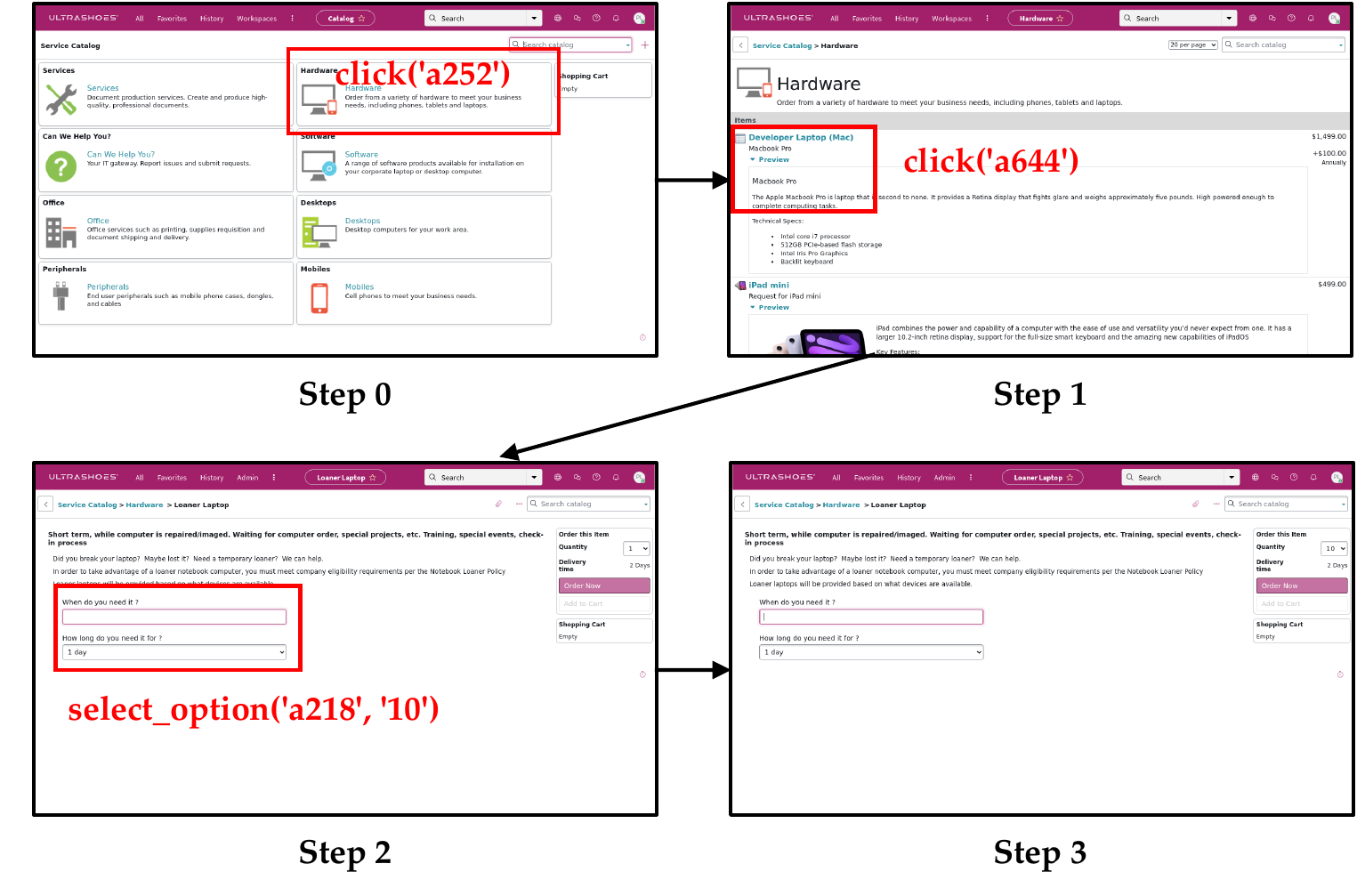}
    \caption{Case study of successful steps 0-3 of \method solving a task of \texttt{Service Catalog}.
    }
  \label{fig:case-steps}
\end{figure*}

Figure~\ref{fig:case-pt1} and Figure~\ref{fig:case-pt2} illustrate a detailed example of \method{} solving a \texttt{List-Sorting} task within WorkArena, where the agent must identify specific features and perform appropriate sorting actions. At each step, \method{} generates a comprehensive reasoning sequence, including target identification, detailed observation analysis, and precise action prediction. These extensive reasoning steps culminate in accurately executed actions, clearly demonstrating \method{}'s effectiveness in systematically navigating complex web interactions.

Figure~\ref{fig:case-steps} provides an additional case study on a \texttt{Service Catalog} task. This example underscores a critical characteristic of web-browsing tasks: each action step significantly transforms the webpage content and corresponding observations (HTML and AXTree structures), rendering multi-step forward planning practically infeasible. The unpredictable and dynamic nature of the observed changes after each interaction highlights the necessity of robust, single-step reasoning capabilities. Consequently, this insight validates our approach of emphasizing single-step decision-making offering valuable guidance for designing future web agent strategies in similarly dynamic environments.

\end{document}